\documentclass{bmvc2k}
\usepackage{graphicx}
\usepackage{booktabs}
\usepackage{amsfonts}
\usepackage{float}
\usepackage{adjustbox}
\usepackage{multirow}

\title{RAIDAL: Redundancy-Aware Information Density Active Learning for CTC-Based Continuous Sign Language Recognition}

\addauthor{\hspace*{8mm}Rafael A. Diniz Augusto}{rada@ufmg.br}{1}
\addauthor{\hspace*{8mm}Gabriel L. Oliveira}{gabriel.leivasoliveira@bristol.ac.uk}{2}
\addauthor{\hspace*{8mm}Erickson R. Nascimento}{erickson@dcc.ufmg.br}{1}

\addinstitution{
Universidade Federal de Minas Gerais\\
Belo Horizonte, Minas Gerais, Brazil
}
\addinstitution{
University of Bristol\\
Bristol, UK
}

\runninghead{Diniz Augusto et al.}{RAIDAL: Active Learning for CSLR}

\def\etal{\emph{et al}\bmvaOneDot}

\begin{document}

\maketitle
\begin{abstract}
Continuous sign language recognition (CSLR) is a key technology for accessibility, yet its development remains limited by the high cost of annotating continuous video streams. Active learning offers a path toward mitigating this cost, but standard acquisition functions are not designed for weakly aligned sign language videos, where sign executions are interleaved with rest poses, irregular pauses, sign-like motion, and temporally redundant frames. This temporal redundancy can undermine sample selection, as acquisition scores may be influenced by timesteps from regions that are not associated with the decoded gloss sequence, distorting the video's estimated informativeness. In this work, we show that modern CSLR models already contain a mechanism for identifying gloss-level temporal evidence: the CTC decoder. Although typically used only during inference, its alignment peaks indicate where the model localizes each predicted gloss in the feature sequence, providing a source of temporal structure for active learning acquisition functions at zero additional labeling cost. Thus, we introduce RAIDAL (Redundancy-Aware Information Density Active Learning), which repurposes the CTC decoder to restrict representation-based scoring to decoder-aligned gloss regions, rather than exposing the acquisition function to the entire unfiltered video. Across three datasets and two architectures, RAIDAL achieves its strongest data-efficiency gains over competing baselines in large-vocabulary, budget-limited settings, while remaining competitive in the smaller-vocabulary, large-budget setting. The code used in this work is publicly available at \href{https://github.com/verlab/RAIDAL}{github.com/verlab/RAIDAL}.
\end{abstract}

%------------------------------------------------------------------------- 
\section{Introduction}
\label{sec:intro}

Sign languages are a primary mode of communication for millions of individuals worldwide, including deaf, hard-of-hearing, and non-speaking people, as well as hearing individuals who regularly interact with sign language users. To bridge the gap between the signing and non-signing communities, continuous sign language recognition (CSLR) aims to automatically transcribe sign language video streams into gloss sequences.

Unlike isolated sign language recognition (ISLR), which classifies pre-segmented clips into single glosses (e.g., ``CAT'', ``HOUSE''), CSLR operates on unsegmented, continuous video data containing full sentences, without access to timestamps for each sign. This distinction introduces a fundamental challenge: the model must learn to align gloss-level annotations to long visual sequences in which signs are interleaved with non-signing content, such as rest poses and irregular pauses, as well as visually similar but non-lexical motion and temporally redundant frames. To address this problem, modern CSLR models are commonly trained with Connectionist Temporal Classification (CTC)~\cite{ctc}, which has become the dominant paradigm for learning from unaligned gloss sequences in CSLR \cite{review2, review1}. CTC enables the model to learn from unaligned features without timestamp supervision by introducing a ``blank'' symbol that the model can emit at any timestep where it is not committing to a gloss. In practice, CTC posteriors are peaky: each gloss is emitted as a sharp activation at one or a few timesteps, with blank predicted almost everywhere else \cite{ctcIsPeaky}. Meanwhile, the CTC decoder is used at inference time to convert the CTC outputs into a predicted gloss sequence.

Despite continued interest in developing new models in this field, annotating continuous sign language videos with gloss labels remains expensive, often requiring dozens of minutes of expert work to label a single minute of video. Duarte \etal report that gloss annotation for How2Sign, a continuous sign language dataset, required, on average, one hour of expert work to annotate 90 seconds of video \cite{annotationcost3}, while Stein \etal report that, after two months of annotation experience, an annotator took around 4 hours to annotate three RWTH-Phoenix weather forecast videos of roughly one minute each \cite{annotationcost4}.

Active learning (AL) offers a path to reduce this cost by selecting only the most informative samples for labeling. Standard AL methods, however, were developed primarily for tasks where the scoring function is not required to reason about weak temporal alignment. In CSLR, this assumption becomes limiting because temporal redundancy can allow the wrong regions of the video to influence the acquisition function: rest poses, irregular pauses, sign-like motion, and high-confidence local activations can affect a sample's uncertainty or diversity estimate even if they were not selected by the CTC decoder as evidence for any predicted gloss.

Our central observation is that CSLR models already include a mechanism that can link predicted glosses to temporal features: the CTC decoder. The decoder is largely treated as an inference-only component in the literature: it is used to obtain the predicted gloss sequence and is then not further utilized. However, the decoding process also produces alignment peaks that can be used to index, at zero additional labeling cost, \emph{the timesteps that the model most strongly associates with each decoded gloss}. Our method, RAIDAL (Redundancy-Aware Information Density Active Learning), recovers the alignment peaks produced by the decoder, uses them to index the feature sequence, and computes acquisition scores from the resulting decoder-aligned features rather than from the full unfiltered video.

A natural alternative is to ignore the CTC decoder and softly weight frame contributions by their CTC probabilities, giving larger influence to frames the model considers non-blank (one minus the CTC blank probability). However, we find that this soft strategy is not sufficient to obtain consistent gains over Random Sampling, as high-confidence activations outside the decoded alignments can still affect the score. RAIDAL instead applies a hard filter. It extracts the decoder's alignment peaks, expands them by a temporal radius, and restricts representation-based scoring to the resulting regions of interest, excluding features outside the decoder-supported regions.

Our main contributions are:
\begin{itemize}
    \item We identify the CTC decoder as a dual-use component of CSLR models: it is traditionally used only at inference time to produce the predicted gloss sequence, but its alignment peaks can also serve as temporal anchors for active learning;
    
    \item We present, to the best of our knowledge, the first study of active learning for continuous sign language recognition, benchmarking uncertainty-based and representation-based strategies across CNN and Transformer architectures, and show that standard active learning methods perform inconsistently in this domain;

    \item We introduce RAIDAL, an active learning acquisition strategy that repurposes the CTC decoder to restrict feature-space diversity scoring to the video's decoder-aligned regions.

\end{itemize}

\section{Related Work}

\subsection{Active Learning for Sign Language}

Prior work in AL for sign language has focused on isolated recognition settings \cite{Fingerspelling-al, IsolatedSign-AL}, where each sample is an independently labeled data point containing a single sign or fingerspelled character. Active learning can treat these settings as standard classification problems, since the model only needs to select which isolated samples are the most useful to label.

However, this formulation does not directly extend to continuous sign language recognition. CSLR operates on unsegmented video streams where multiple gloss executions are interleaved with rest poses, transitions, and temporally redundant content, while supervision is provided only as the target gloss sequence. As a result, the acquisition function must decide which videos are worth annotating without access to frame-level sign boundaries, and without assuming that every temporal region contributes equally to the target sequence.

To the best of our knowledge, no prior work has addressed active learning for continuous sign language recognition, leaving open the question of how standard acquisition strategies behave in weakly aligned sign language video streams.

\subsection{Active Learning for Sequence-Level Tasks}

When applying active learning to sequential domains, most standard acquisition functions require temporal aggregation to provide a single sequence-level score.

For uncertainty-based selection, mean entropy is a common adaptation in the literature. By computing the entropy of each temporal segment and then taking the average, this strategy serves as a sequence-level uncertainty score. Recent works across different sequential domains \cite{meanEntropy1, meanEntropy2} explicitly use this average entropy of segments as a sequence-level acquisition strategy or baseline. Additionally, recent work such as UNCAST \cite{UNCAST} proposes measuring uncertainty directly at the sequence level instead of operating on isolated segments.

For representation-based selection, Core-set \cite{coreset} remains a widely used strategy. Since the original method does not specify how to represent a sequence of features as a single instance, a simple way to obtain such an instance-level representation is to average features along the sequence dimension into a single vector, a technique used more broadly when encoding sequential inputs \cite{averagePooling1, averagePooling2}. This same averaging strategy has also been used in the literature to construct the feature vector on which Core-set operates \cite{pooledCoreset}.

While these adaptations enable sequence-level active learning, they do not reliably transfer to CSLR. Uncertainty-based methods can increase vocabulary coverage while leaving many selected glosses with few examples, global representation methods can miss execution-level variation by collapsing time, and frame-level representation methods can still allow timesteps outside decoder-supported regions to influence the sample's score. RAIDAL addresses this gap by repurposing the decoder's alignment peaks as temporal anchors, restricting representation-based selection to decoder-aligned gloss regions.

\section{Methodology}

\subsection{Problem Formulation}
We consider the standard pool-based active learning setup for continuous sign language recognition. Let $\mathcal{U} = \{x_i\}_{i=1}^{N}$ be a pool of unlabeled videos, where $x_i \in \mathbb{R}^{T_i \times H \times W \times C}$ is a video sequence of length $T_i$. Our goal is to iteratively select a batch of samples $\mathcal{B} \subset \mathcal{U}$ to be annotated by an oracle and added to the labeled training set $\mathcal{L}$, such that the model performance is maximized under a fixed labeling budget $b$.

Unlike in standard classification, where the label $y_i$ is a single class index, the target $y_i = (w_1, \dots, w_L)$ in CSLR is a sequence of glosses of length $L$, where $L \ll T_i$. The lack of frame-level alignment between $x_i$ and $y_i$ motivates the use of a CTC-based model $f_\theta$, which maps the input video to a sequence of class probability distributions.

\subsection{RAIDAL: Redundancy-Aware Information Density Active Learning}

\begin{figure}[t!]
    \centering
    \includegraphics[width=\linewidth]{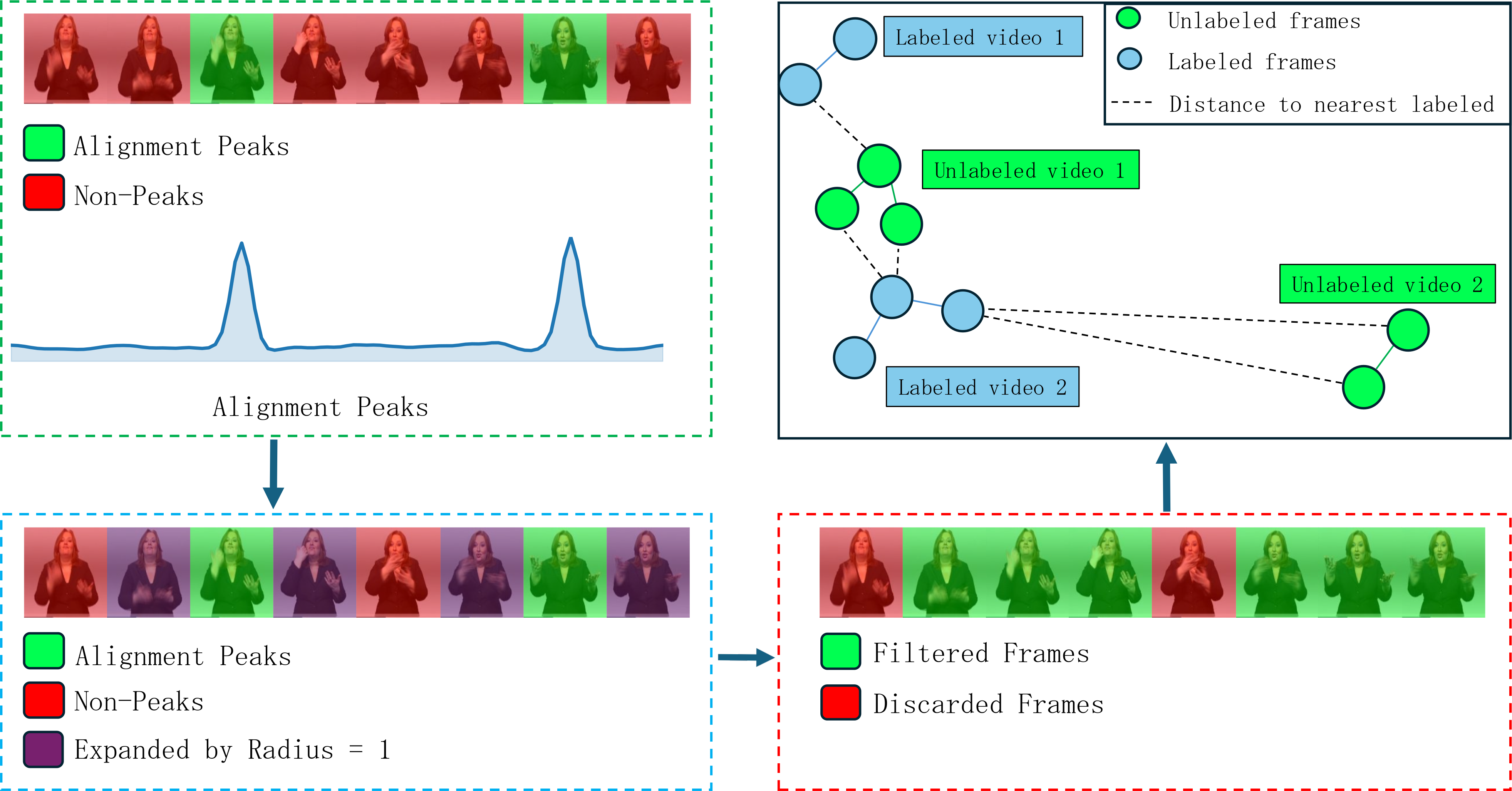}
    \caption{\textbf{Overview of RAIDAL}. Alignment peaks are expanded by a temporal radius to retain local context around the decoder-aligned frames. Candidate videos are then scored by the distance between their filtered features and the labeled pool, normalized by raw video length to favor samples with high useful diversity relative to annotation cost.}
    \label{fig:raidal_pipeline}
\end{figure}

RAIDAL repurposes the CTC decoder to identify the timesteps most associated with decoded glosses, and then measures representation diversity only over these decoder-aligned regions. The method has two stages: i) Redundancy-Aware Filtering and ii) Information Density. Figure~\ref{fig:raidal_pipeline} provides a visual overview of RAIDAL.

\subsubsection{Redundancy-Aware Filtering}
\label{raidal: redundancy aware filtering}

The CSLR models used in our experiments use CTC beam search decoding in their public implementations \cite{CorrNetRepo, SwinMSTPRepo}. RAIDAL reuses this existing decoder and extracts temporal anchors from its decoded hypotheses. Unlike greedy decoding, which considers each frame's label independently, beam search scores hypotheses over the whole sequence, so the glosses it selects form a hypothesis that is consistent at the sequence level rather than chosen frame by frame. For an unlabeled video $x$, the decoder returns the top-$K$ decoded hypotheses $\mathcal{H}$. For each hypothesis $h \in \mathcal{H}$, we extract a set of alignment peaks $P_h = \{p_1, \dots, p_{n}\}$, where $p_i$ is the timestep of maximum activation for the $i$-th decoded gloss. RAIDAL uses these peaks as temporal anchors, retaining features that are associated with at least one decoded path.

Since a single peak may not cover the full physical execution of a sign, we introduce a Temporal Expansion Radius ($R$). We define $R$ as a function of the video's frame rate ($\text{FPS}$), the model's temporal downsampling stride ($S$), and the sign duration in seconds ($T_{sign}$):

\begin{equation}
    \label{eq:temporal_expansion}
    R = \left\lfloor \frac{1}{2} \left( \frac{T_{sign} \times \text{FPS}}{S} - 1 \right) \right\rceil.
\end{equation}

For the temporal expansion, we use $T_{sign}=0.5$ seconds as an estimate of sign duration. This value is consistent with sign durations reported for multiple sign languages \cite{avgSignDurationASL, avgSignDurationEurope, avgSignDurationCSL}. For each peak $p \in P_h$, we define a temporal window $W(p)$ centered on the peak:

\begin{equation}
    \label{eq:window}
    W(p) = \{ t \in \mathbb{N} \mid p - R \le t \le p + R \}.
\end{equation}

For instance, for a video recorded at $\text{FPS}=25$ and a model with temporal downsampling stride $S=4$, the Temporal Expansion Radius formula (Equation \ref{eq:temporal_expansion}) yields $R=1$. Using $R=1$ with Equation \ref{eq:window} creates a 3-timestep window that gives local context to each alignment peak. Note that if $R$ is a decimal number, it is rounded to the nearest integer by Equation \ref{eq:temporal_expansion}.

To account for different alignment peaks across hypotheses, RAIDAL aggregates the expanded peak windows from all hypotheses to form a Temporal Region of Interest $\mathcal{TROI}$:

\begin{equation}
    \mathcal{TROI} = \bigcup_{h \in \mathcal{H}} \bigcup_{p \in P_h} W(p).
\end{equation}

We use the $\mathcal{TROI}$ to slice the dense video features $F$, producing a filtered feature sequence $\hat{F} = \{f_t \in F \mid t \in \mathcal{TROI}\}$. This provides a structural alternative to manually tuned confidence thresholds, leveraging the model's decoding structure to retain only features supported by at least one of the decoded hypotheses $\mathcal{H}$, and preventing timesteps outside the $\mathcal{TROI}$ from influencing the acquisition score.

We highlight that it is important to distinguish decoder-supported regions from the physical boundaries of sign executions. Alignment peaks are derived from the model's own decoded hypotheses and should therefore be interpreted as \emph{the timesteps the model uses to support its current predictions}, rather than as direct detections of the glosses contained in the video. Because the CSLR datasets considered in this work provide only sequence-level gloss annotations without frame-level temporal boundaries, we cannot quantitatively verify how closely these peaks correspond to the actual sign executions. We therefore interpret them strictly as model-derived temporal anchors for acquisition.

\subsubsection{Information Density}
\label{raidal: information density}

After filtering, RAIDAL scores each candidate video by the distance between its filtered features and the labeled pool. Let $\mathcal{L}$ be the set of labeled videos, $F_{\mathcal{L}}$ the set of features extracted from labeled videos, and $\mathcal{U}$ the pool of unlabeled videos. For a candidate video $u \in \mathcal{U}$ with raw frame length $T_u$, we define its distance to the labeled pool as the sum of the distances from each filtered feature $f_u \in \hat{F}_u$ to its nearest labeled feature $f_l \in F_{\mathcal{L}}$, normalized by $T_u$. The purpose of this score is therefore not to measure diversity over the entire video, but to measure how much new feature-space novelty relative to the labeled pool is present in the regions the decoder selected as gloss evidence. Note that our use of Information Density refers to filtered feature variation per unit of annotation cost, which differs from Settles's original terminology, in which an informativeness term is weighted by a representativeness term \cite{settlesActiveLearning}:

\begin{equation}
    \label{eq:RaidalScoreFunc}
    \Delta(u, \mathcal{L}) = \frac{1}{T_u} \sum_{f_u \in \hat{F}_u} \min_{f_l \in F_{\mathcal{L}}} \| f_u - f_l \|_2.
\end{equation}

We then use a greedy selection procedure following the Core-set principle of iteratively expanding feature-space coverage~\cite{coreset}: at each iteration, the candidate $u^* = \arg\max_{u \in \mathcal{U}} \Delta(u, \mathcal{L})$ is selected and its filtered features $\hat{F}_{u^*}$ are added to $F_{\mathcal{L}}$ to avoid resampling videos with similar features. This procedure repeats until the frame budget is exhausted.

We stress that the normalization by the raw video length $T_u$ is intentional. If the score were normalized by the length of $\hat{F}_u$, long videos with only a few distant filtered timesteps could receive inflated scores, even when most of their annotation cost comes from unselected regions. Normalizing by $T_u$ instead favors videos whose distance from the labeled feature set is high relative to their total annotation cost. 

\section{Experiments}

\subsection{Experimental Setup}

\textbf{Datasets.} We evaluate our method on three CSLR datasets: PHOENIX-2014 \cite{phoenix2014}, PHOENIX-T \cite{phoenix-T}, and Isharah-1000 \cite{Isharah}. All datasets provide continuous sign language videos with gloss-level annotations, and overall performance is measured using the word error rate (WER). PHOENIX-2014 and PHOENIX-T are two of the most widely used CSLR benchmarks. Both have large vocabularies of $1{,}231$ and $1{,}085$ unique glosses, respectively, and are recorded in a controlled environment. In contrast, Isharah-1000 is a newly released dataset featuring unconstrained footage. Its footage was captured using smartphone cameras across diverse environments, introducing multiple sources of variation, including varying resolutions, camera angles, and lighting conditions. The dataset also has a comparatively smaller vocabulary of $680$ unique glosses. Evaluating RAIDAL on Isharah-1000 serves as a useful proxy for measuring how our method performs under more realistic visual conditions, although its smaller vocabulary means that it may not capture the full lexical complexity of real-world CSLR, where vocabularies can contain thousands of unique signs.

\noindent\textbf{Architectures and Training.} To evaluate whether RAIDAL transfers across different backbone architectures, we test it on both CNN and Transformer models. Our primary architecture is CorrNet \cite{corrnet}, a CNN-based model, and we additionally test RAIDAL on Swin-MSTP \cite{swin_mstp}, a Transformer-based model.

We adopt a pool-based setup with the annotation budget measured in total frames to ensure a fair comparison across videos of varying lengths. For the PHOENIX-2014 and PHOENIX-T datasets, all methods start from the same initial labeled pool of $2.5\%$ of the total available frames (20k and 20.7k frames, respectively). The models then query batches following a progressive percentage-based schedule up to $25.0\%$ (yielding maximum budgets of 200k and 207k frames). Conversely, due to the much higher data requirements for testing on the Isharah dataset, we evaluate it using a larger, fixed schedule starting at a shared initial pool of 200k frames, progressing in 100k increments up to 700k frames.

To account for statistical variance, we report the WER mean and standard deviation over three independent replications in all tests, analyses, and ablations. We use identical random seeds across all methods in each run (i.e., the \emph{i}-th replication uses the same seed for every method and every budget), reducing confounding from initialization variance.

\subsection{Baselines}

We compare RAIDAL against temporal adaptations of AL methods found in the literature. We also introduce a domain-specific frame-level baseline to compare against our method.

\subsubsection{Representation-Based Strategies}

\begin{itemize}
\item \textbf{Core-set:} We adopt the temporal adaptation of Core-set used in prior work \cite{pooledCoreset}. Each video is represented by a single globally pooled embedding, and acquisition maximizes feature-space distances in order to promote diversity in the labeled pool.

\item \textbf{Soft-Attention Core-set:} Our own domain-adapted frame-level representation baseline. Soft-Attention Core-set calculates feature distances between individual frames and softly weights the contribution of each frame by its non-blank probability.

Note that this variant is not a method previously published in the literature. We designed it as a controlled baseline that provides a direct mechanical contrast to RAIDAL, isolating the performance gap between retaining all frames with soft probability weighting and applying a hard filter based on the CTC decoder.

\end{itemize}

\subsubsection{Uncertainty-Based Strategies}
\begin{itemize}
\item \textbf{Entropy:} We follow sequence-level adaptations of Entropy Sampling used in prior work \cite{meanEntropy1, meanEntropy2}. For each video, the acquisition score is computed by averaging the class entropy across all frames.

\item \textbf{UNCAST~\cite{UNCAST}:} Although originally proposed for self-training, we adapt the sequence-level CTC uncertainty estimator introduced in UNCAST as an acquisition score. Unlike Entropy, it estimates uncertainty directly at the decoded-sequence level using the CTC loss and Monte Carlo dropout.

\end{itemize}

\subsubsection{Gradient-Based Strategies}

\begin{itemize}
    \item \textbf{CTC-BADGE:} We adapt BADGE \cite{badge} to CSLR by computing gradient embeddings from the CTC loss with the top decoded hypothesis as a pseudo-label, followed by k-means++ batch selection. To account for variable video lengths, we normalize summed gradient contributions by the number of frames. We evaluate this additional baseline on our primary PHOENIX-2014/CorrNet setting.
\end{itemize}

\section{Results}
\label{sec:results}

\subsection{Main Comparison on PHOENIX-2014 and PHOENIX-T}
\label{subsec:corrnet}

We first evaluate RAIDAL on PHOENIX-2014 and PHOENIX-T using CorrNet, our main architecture. Tables \ref{tab:wer_phoenix14} and \ref{tab:wer_phoenixT} report the WER progression across annotation budgets (lower WER is better), the normalized area under the learning curve (nAULC, lower is better, computed as the area under the WER-versus-budget curve divided by the budget range), and $p$-values from paired $t$-tests comparing each method's nAULC with RAIDAL's.

On PHOENIX-2014, RAIDAL obtains the lowest nAULC and significantly outperforms Random Sampling, Entropy, Core-set, UNCAST, Soft-Attention Core-set, and CTC-BADGE. On PHOENIX-T, RAIDAL again achieves the lowest nAULC, with significant gains over Entropy, Core-set, UNCAST, and Soft-Attention Core-set, while its improvement over Random Sampling is not statistically significant ($p=0.064$). Notably, Entropy, Core-set, and UNCAST achieve worse nAULC than Random Sampling on both datasets, suggesting that standard active learning strategies do not reliably transfer to CSLR.

The composition of the labeled pools (Tables \ref{tab:supp_p14_corrnet} and \ref{tab:supp_pt_corrnet}) offers a possible explanation for RAIDAL's stronger performance. Entropy, UNCAST, and Core-set generally cover a broader vocabulary than RAIDAL, but this broader coverage comes with more singleton glosses and lower mean frequencies per gloss. For Entropy and UNCAST, we argue that rare or previously unseen glosses tend to induce higher uncertainty, leading these methods to allocate more of the annotation budget toward vocabulary breadth. When many selected glosses have only a few training examples, these broader but shallower labeled pools may provide insufficient repeated exposure for learning robust visual representations, ultimately resulting in worse WER despite their greater vocabulary diversity. In contrast, RAIDAL constructs a narrower but deeper labeled pool, with slightly lower vocabulary coverage but a substantially higher mean per-gloss frequency. This balance between vocabulary coverage and sufficient per-gloss representation may help explain RAIDAL's stronger data efficiency and indicates that vocabulary coverage alone may not be a reliable proxy for sample utility.

Another key finding comes from Soft-Attention Core-set. This baseline provides the closest comparison to RAIDAL, as it also operates at the timestep level and avoids treating all timesteps equally by weighting each timestep's distance contribution by its corresponding non-blank probability. Soft-Attention Core-set additionally presents a vocabulary profile that is very close to RAIDAL's, despite obtaining substantially worse WER. Its worse performance despite constructing a similarly distributed labeled pool suggests that soft weighting alone may not be sufficient, and that explicitly restricting which timesteps are allowed to influence the acquisition score can provide additional value beyond the vocabulary composition of the selected pool.

\begin{table}[t]
    \centering
    \small
    \setlength{\tabcolsep}{3pt}
    \resizebox{\linewidth}{!}{
    \begin{tabular}{lccccccccccc}
        \toprule
        \multicolumn{12}{c}{\textbf{PHOENIX-2014 (CorrNet)}} \\
        \midrule
        \multirow{2}{*}{\textbf{Method}} & \multicolumn{9}{c}{\textbf{Annotation Budget in Frames}} & \multirow{2}{*}{\textbf{nAULC}} & \multirow{2}{*}{\textbf{$p$-value}} \\
        \cmidrule{2-10}
         & \textbf{30k} & \textbf{40k} & \textbf{50k} & \textbf{60k} & \textbf{80k} & \textbf{100k} & \textbf{120k} & \textbf{160k} & \textbf{200k} & & \\
        \midrule
        Random & $76.3_{\pm0.9}$ & $65.7_{\pm0.4}$ & $59.4_{\pm1.6}$ & $53.5_{\pm1.2}$ & $47.2_{\pm0.7}$ & $42.3_{\pm0.6}$ & $39.5_{\pm0.4}$ & $34.3_{\pm0.4}$ & $30.9_{\pm0.3}$ & $45.5_{\pm0.6}$ & $0.031$ \\
        \midrule
        Entropy & $76.1_{\pm0.4}$ & $66.2_{\pm0.4}$ & $58.2_{\pm1.2}$ & $55.9_{\pm1.8}$ & $47.6_{\pm0.9}$ & $42.3_{\pm0.2}$ & $39.7_{\pm0.5}$ & $34.4_{\pm0.6}$ & $30.4_{\pm0.3}$ & $45.8_{\pm0.2}$ & $<0.001$ \\
        \midrule
        UNCAST & $78.4_{\pm0.1}$ & $72.3_{\pm1.4}$ & $63.2_{\pm0.9}$ & $59.2_{\pm0.5}$ & $50.4_{\pm1.1}$ & $46.7_{\pm1.0}$ & $41.9_{\pm0.5}$ & $35.5_{\pm0.4}$ & $31.9_{\pm0.6}$ & $48.4_{\pm0.4}$ & $0.005$ \\
        \midrule
        Core-set & $75.1_{\pm0.8}$ & $66.1_{\pm1.8}$ & $59.0_{\pm0.7}$ & $53.5_{\pm1.0}$ & $48.2_{\pm0.6}$ & $43.5_{\pm1.0}$ & $39.7_{\pm0.7}$ & $34.5_{\pm0.1}$ & $31.1_{\pm0.2}$ & $45.8_{\pm0.3}$ & $0.013$ \\
        \midrule
        Soft-Attn Core-set & $72.7_{\pm0.9}$ & $64.0_{\pm0.3}$ & $56.7_{\pm0.5}$ & $53.6_{\pm0.5}$ & $46.9_{\pm0.2}$ & $42.5_{\pm0.5}$ & $38.6_{\pm0.1}$ & $34.7_{\pm0.3}$ & $31.2_{\pm0.3}$ & $45.1_{\pm0.1}$ & $0.002$ \\
        \midrule
        CTC-BADGE & $74.7_{\pm0.6}$ & $65.2_{\pm1.3}$ & $57.8_{\pm0.9}$ & $53.8_{\pm0.9}$ & $47.6_{\pm0.5}$ & $42.4_{\pm0.1}$ & $38.7_{\pm0.7}$ & $34.0_{\pm0.3}$ & $30.9_{\pm0.4}$ & $45.2_{\pm0.4}$ & $0.037$ \\
        \midrule
        \textbf{RAIDAL (Ours)} & $\mathbf{71.8_{\pm0.5}}$ & $\mathbf{61.2_{\pm1.6}}$ & $\mathbf{54.9_{\pm1.0}}$ & $\mathbf{51.3_{\pm0.6}}$ & $\mathbf{44.3_{\pm0.2}}$ & $\mathbf{40.5_{\pm0.4}}$ & $\mathbf{37.2_{\pm0.4}}$ & $\mathbf{33.1_{\pm0.2}}$ & $\mathbf{29.8_{\pm0.6}}$ & $\mathbf{43.3_{\pm0.2}}$ & -- \\
        \bottomrule
    \end{tabular}
    }
    \caption{WER (\%) and nAULC on PHOENIX-2014 using CorrNet. Results are mean $\pm$ std over 3 runs; $p$-values are obtained from $t$-tests against RAIDAL. The initial 20k-frame budget is omitted because it yields a shared baseline performance of 83.26\% $\pm$ 0.73.}
    \label{tab:wer_phoenix14}
\end{table}

\begin{table}[t]
    \centering
    \small
    \setlength{\tabcolsep}{3pt}
    \resizebox{\linewidth}{!}{
    \begin{tabular}{lccccccccccc}
        \toprule
        \multicolumn{12}{c}{\textbf{PHOENIX-T (CorrNet)}} \\
        \midrule
        \multirow{2}{*}{\textbf{Method}} & \multicolumn{9}{c}{\textbf{Annotation Budget in Frames}} & \multirow{2}{*}{\textbf{nAULC}} & \multirow{2}{*}{\textbf{$p$-value}} \\
        \cmidrule{2-10}
         & \textbf{31k} & \textbf{41k} & \textbf{52k} & \textbf{62k} & \textbf{82k} & \textbf{103k} & \textbf{124k} & \textbf{165k} & \textbf{207k} & & \\
        \midrule
        Random & $79.6_{\pm1.1}$ & $70.4_{\pm1.5}$ & $\mathbf{58.9_{\pm1.0}}$ & $54.2_{\pm0.7}$ & $46.2_{\pm1.1}$ & $40.8_{\pm0.5}$ & $38.2_{\pm0.1}$ & $34.0_{\pm0.8}$ & $30.7_{\pm0.4}$ & $45.6_{\pm0.5}$ & $0.064$ \\
        \midrule
        Entropy & $77.5_{\pm1.7}$ & $69.3_{\pm1.8}$ & $61.4_{\pm0.4}$ & $56.6_{\pm0.6}$ & $48.3_{\pm1.2}$ & $43.6_{\pm0.8}$ & $38.7_{\pm0.4}$ & $33.6_{\pm0.1}$ & $30.1_{\pm0.3}$ & $46.3_{\pm0.3}$ & $0.021$ \\
        \midrule
        UNCAST & $84.2_{\pm1.0}$ & $72.0_{\pm1.1}$ & $66.7_{\pm0.9}$ & $58.5_{\pm1.6}$ & $50.3_{\pm0.9}$ & $43.9_{\pm0.8}$ & $40.7_{\pm1.3}$ & $35.6_{\pm1.3}$ & $31.3_{\pm0.3}$ & $48.4_{\pm0.6}$ & $0.015$ \\
        \midrule
        Core-set & $78.4_{\pm2.3}$ & $72.1_{\pm2.4}$ & $62.9_{\pm0.8}$ & $56.6_{\pm1.0}$ & $49.6_{\pm0.9}$ & $44.6_{\pm0.6}$ & $39.8_{\pm0.6}$ & $35.1_{\pm0.5}$ & $30.8_{\pm0.2}$ & $47.4_{\pm0.2}$ & $0.005$ \\
        \midrule
        Soft-Attn Core-set & $77.7_{\pm1.6}$ & $69.5_{\pm1.1}$ & $61.6_{\pm2.0}$ & $55.6_{\pm1.6}$ & $48.6_{\pm1.1}$ & $43.0_{\pm0.4}$ & $38.1_{\pm0.2}$ & $33.7_{\pm0.5}$ & $30.4_{\pm0.4}$ & $46.1_{\pm0.5}$ & $0.033$ \\
        \midrule
        \textbf{RAIDAL (Ours)} & $\mathbf{75.4_{\pm1.3}}$ & $\mathbf{66.4_{\pm1.3}}$ & $59.4_{\pm1.0}$ & $\mathbf{53.6_{\pm1.4}}$ & $\mathbf{43.5_{\pm0.2}}$ & $\mathbf{39.0_{\pm0.8}}$ & $\mathbf{36.1_{\pm0.5}}$ & $\mathbf{32.1_{\pm0.6}}$ & $\mathbf{29.4_{\pm0.2}}$ & $\mathbf{43.7_{\pm0.3}}$ & -- \\
        \bottomrule
    \end{tabular}
    }
    \caption{WER (\%) and nAULC on PHOENIX-T using CorrNet. Results are mean $\pm$ std over 3 runs; $p$-values are obtained from $t$-tests against RAIDAL. The initial 20.7k-frame budget is omitted because it yields a shared baseline performance of 90.14\% $\pm$ 1.06.}
    \label{tab:wer_phoenixT}
\end{table}

\subsection{Results on Swin-MSTP}

To assess whether RAIDAL transfers beyond the CorrNet model \cite{corrnet}, we evaluate it using the Transformer-based Swin-MSTP model \cite{swin_mstp}. Because performance begins to converge at the 160k and 165k budgets, we stop the labeling process at these points. Tables~\ref{tab:wer_swin_phoenix14} and~\ref{tab:wer_swin_phoenixT} show that RAIDAL obtains the lowest nAULC on both PHOENIX datasets, with statistically significant differences from Random, Entropy, Core-set, and UNCAST. Soft-Attention Core-set is the only baseline for which the comparison is not statistically significant on PHOENIX-2014 ($p=0.088$), despite being significant on PHOENIX-T ($p=0.013$). 

The labeled pools also follow a similar pattern to the CorrNet experiments. Entropy, Core-set, and UNCAST again acquire wider vocabularies with fewer examples per gloss, while RAIDAL constructs a narrower but deeper labeled pool (Tables \ref{tab:supp_phoenix14_swin} and \ref{tab:supp_pt_swin}). Soft-Attention Core-set again remains close to RAIDAL at the vocabulary level, further suggesting that vocabulary composition alone does not fully explain the difference in performance between these two methods. Overall, the results show that RAIDAL transfers to the Transformer-based Swin-MSTP, although the magnitude of its advantage over Soft-Attention Core-set is not uniform across the evaluated settings.

\begin{table}[t]
    \centering
    \small
    \setlength{\tabcolsep}{3pt}
    \resizebox{\linewidth}{!}{
    \begin{tabular}{lcccccccccc}
        \toprule
        \multicolumn{11}{c}{\textbf{PHOENIX-2014 (Swin-MSTP)}} \\
        \midrule
        \multirow{2}{*}{\textbf{Method}} & \multicolumn{8}{c}{\textbf{Annotation Budget in Frames}} & \multirow{2}{*}{\textbf{nAULC}} & \multirow{2}{*}{\textbf{$p$-value}} \\
        \cmidrule{2-9}
         & \textbf{30k} & \textbf{40k} & \textbf{50k} & \textbf{60k} & \textbf{80k} & \textbf{100k} & \textbf{120k} & \textbf{160k} & & \\
        \midrule
        Random & $73.9_{\pm1.0}$ & $64.2_{\pm1.0}$ & $55.2_{\pm0.9}$ & $50.3_{\pm0.5}$ & $43.4_{\pm0.3}$ & $38.7_{\pm0.5}$ & $35.4_{\pm0.2}$ & $31.5_{\pm0.1}$ & $45.9_{\pm0.4}$ & $0.019$ \\
        \midrule
        Entropy & $74.3_{\pm0.6}$ & $65.6_{\pm1.4}$ & $57.3_{\pm1.1}$ & $50.7_{\pm0.1}$ & $43.5_{\pm0.5}$ & $38.8_{\pm0.7}$ & $35.3_{\pm0.2}$ & $31.0_{\pm0.4}$ & $46.2_{\pm0.1}$ & $0.006$ \\
        \midrule
        UNCAST & $77.0_{\pm0.6}$ & $69.4_{\pm0.4}$ & $60.3_{\pm2.0}$ & $53.8_{\pm0.2}$ & $46.4_{\pm0.2}$ & $40.9_{\pm0.5}$ & $37.2_{\pm0.3}$ & $31.8_{\pm0.2}$ & $48.5_{\pm0.1}$ & $0.001$ \\
        \midrule
        Core-set & $76.0_{\pm0.4}$ & $66.6_{\pm0.4}$ & $57.0_{\pm0.9}$ & $51.5_{\pm0.5}$ & $43.8_{\pm0.3}$ & $39.0_{\pm0.3}$ & $36.3_{\pm0.1}$ & $32.1_{\pm0.3}$ & $46.9_{\pm0.2}$ & $0.001$ \\
        \midrule
        Soft-Attn Core-set & $72.7_{\pm1.2}$ & $\mathbf{62.0_{\pm0.6}}$ & $54.6_{\pm1.0}$ & $49.3_{\pm0.8}$ & $42.9_{\pm0.6}$ & $37.8_{\pm0.6}$ & $35.1_{\pm0.6}$ & $31.3_{\pm0.3}$ & $45.2_{\pm0.5}$ & $0.088$ \\
        \midrule
        \textbf{RAIDAL (Ours)} & $\mathbf{72.7_{\pm0.2}}$ & $62.7_{\pm0.9}$ & $\mathbf{53.0_{\pm0.7}}$ & $\mathbf{47.3_{\pm0.2}}$ & $\mathbf{41.1_{\pm0.4}}$ & $\mathbf{37.1_{\pm0.2}}$ & $\mathbf{34.2_{\pm0.2}}$ & $\mathbf{30.3_{\pm0.2}}$ & $\mathbf{44.3_{\pm0.1}}$ & -- \\
        \bottomrule
    \end{tabular}
    }
    \caption{WER (\%) and nAULC on PHOENIX-2014 using Swin-MSTP. Results are mean $\pm$ std over 3 runs; $p$-values are obtained from $t$-tests against RAIDAL. The initial 20k-frame budget is omitted because it yields a shared baseline performance of 82.10\% $\pm$ 0.66.}
    \label{tab:wer_swin_phoenix14}
\end{table}

\begin{table}[t]
    \centering
    \small
    \setlength{\tabcolsep}{3pt}
    \resizebox{\linewidth}{!}{
    \begin{tabular}{lcccccccccc}
        \toprule
        \multicolumn{11}{c}{\textbf{PHOENIX-T (Swin-MSTP)}} \\
        \midrule
        \multirow{2}{*}{\textbf{Method}} & \multicolumn{8}{c}{\textbf{Annotation Budget in Frames}} & \multirow{2}{*}{\textbf{nAULC}} & \multirow{2}{*}{\textbf{$p$-value}} \\
        \cmidrule{2-9}
         & \textbf{31k} & \textbf{41k} & \textbf{52k} & \textbf{62k} & \textbf{82k} & \textbf{103k} & \textbf{124k} & \textbf{165k} & & \\
        \midrule
        Random & $73.2_{\pm1.4}$ & $63.7_{\pm0.9}$ & $56.4_{\pm1.0}$ & $50.7_{\pm1.1}$ & $42.4_{\pm0.2}$ & $38.2_{\pm0.7}$ & $35.1_{\pm0.5}$ & $31.1_{\pm0.4}$ & $45.7_{\pm0.5}$ & $0.031$ \\
        \midrule
        Entropy & $74.9_{\pm2.1}$ & $65.7_{\pm1.1}$ & $57.4_{\pm0.9}$ & $50.3_{\pm0.7}$ & $41.9_{\pm0.8}$ & $37.8_{\pm0.4}$ & $34.1_{\pm0.5}$ & $30.3_{\pm0.2}$ & $45.6_{\pm0.2}$ & $0.015$ \\
        \midrule
        UNCAST & $77.2_{\pm1.6}$ & $68.1_{\pm2.0}$ & $60.2_{\pm1.7}$ & $54.1_{\pm1.3}$ & $45.3_{\pm0.6}$ & $41.3_{\pm0.4}$ & $37.2_{\pm0.6}$ & $31.2_{\pm0.3}$ & $48.3_{\pm0.6}$ & $0.009$ \\
        \midrule
        Core-set & $76.0_{\pm0.9}$ & $66.6_{\pm0.4}$ & $58.2_{\pm0.7}$ & $52.0_{\pm0.9}$ & $44.4_{\pm0.7}$ & $39.2_{\pm0.3}$ & $36.5_{\pm0.1}$ & $31.3_{\pm0.3}$ & $47.2_{\pm0.1}$ & $0.007$ \\
        \midrule
        Soft-Attn Core-set & $73.2_{\pm1.7}$ & $62.6_{\pm0.7}$ & $55.1_{\pm1.1}$ & $51.2_{\pm1.7}$ & $43.4_{\pm0.6}$ & $37.5_{\pm0.4}$ & $34.8_{\pm0.2}$ & $31.4_{\pm0.2}$ & $45.7_{\pm0.2}$ & $0.013$ \\
        \midrule
        \textbf{RAIDAL (Ours)} & $\mathbf{71.4_{\pm1.8}}$ & $\mathbf{62.3_{\pm2.1}}$ & $\mathbf{53.2_{\pm0.7}}$ & $\mathbf{47.8_{\pm0.5}}$ & $\mathbf{40.8_{\pm0.2}}$ & $\mathbf{35.9_{\pm0.3}}$ & $\mathbf{33.1_{\pm0.5}}$ & $\mathbf{29.6_{\pm0.4}}$ & $\mathbf{43.8_{\pm0.3}}$ & -- \\
        \bottomrule
    \end{tabular}
    }
    \caption{WER (\%) and nAULC on PHOENIX-T using Swin-MSTP. Results are mean $\pm$ std over 3 runs; $p$-values are obtained from $t$-tests against RAIDAL. The initial 20.7k-frame budget is omitted because it yields a shared baseline performance of 84.91\% $\pm$ 0.29.}
    \label{tab:wer_swin_phoenixT}
\end{table}

\subsection{Results on Isharah}
\label{subsec:Isharah}

Due to its unconstrained footage, Isharah produces a setting of high visual variance, including shifts in camera position, resolution, and environment. Following observations by the Isharah authors that CorrNet and several other evaluated models struggle to perform reliably in low-data regimes on this dataset \cite{Isharah}, we conduct all experiments using the Swin-MSTP architecture.

Table \ref{tab:wer_isharah_swin} shows RAIDAL achieving the lowest nAULC on Isharah at 63.4, significantly outperforming Random, Core-set, and UNCAST (all $p < 0.05$), while showing no significant difference from Entropy ($p=0.703$) and Soft-Attention Core-set ($p=0.470$). The similar performance of RAIDAL, Entropy, and Soft-Attention Core-set is notable given the larger gaps observed on the PHOENIX datasets, and can be partially understood by returning to the breadth-versus-depth mechanism discussed in Section \ref{subsec:corrnet}.

In the PHOENIX datasets, the larger vocabularies and stricter annotation budgets make the allocation of examples across glosses especially important. Uncertainty-based methods tend to acquire broad vocabularies with many low-frequency glosses, while RAIDAL favors feature-space diversity within decoder-supported regions, producing deeper labeled pools with more repeated exposure to selected glosses. Isharah changes this balance because its vocabulary is smaller and its absolute annotation budgets are substantially larger, with all methods already covering more than 90\% of the dataset's $680$ glosses by 300k frames (Table \ref{tab:supp_isharah_swin}). Once most glosses are represented, Entropy has less opportunity to prioritize rare or unseen glosses and increasingly acquires repeated executions, bringing its vocabulary coverage and mean per-gloss frequency close to those of RAIDAL and providing a plausible explanation for why their performance gap narrows.

Soft-Attention Core-set also becomes competitive on Isharah, although its convergence with RAIDAL cannot be fully explained by vocabulary coverage in the same way as Entropy. Soft-Attention Core-set provides a soft counterpart to RAIDAL, sharing its timestep-level, representation-based formulation but weighting the distance contribution of each timestep by its non-blank probability instead of restricting the score to decoder-supported regions. One possible explanation for the small gap in this setting is Isharah's substantially larger absolute budgets and early vocabulary saturation, which may reduce the impact of Soft-Attention's less selective weighting by allowing it to accumulate sufficient useful feature variation despite retaining all timesteps in the score. Although we cannot isolate this effect from the other differences between Isharah and PHOENIX, this interpretation is consistent with the smaller gap observed in this setting.

Overall, across the evaluated benchmarks, RAIDAL's clearest advantage is observed in the larger-vocabulary, more budget-constrained PHOENIX settings, while the gap narrows on the smaller-vocabulary, larger-budget Isharah-1000 setting. Because Isharah simultaneously differs from PHOENIX in vocabulary size and visual variation, we cannot decisively determine whether RAIDAL's advantage increases with vocabulary size, although the observed vocabulary patterns are consistent with this hypothesis. We revisit this confound in Section~\ref{subsec:futurework}, where we outline a more direct test of the role of vocabulary size.

\begin{table}[t]
    \centering
    \small
    \resizebox{\linewidth}{!}{
    \begin{tabular}{lccccccc}
        \toprule
        \multicolumn{8}{c}{\textbf{Isharah (Swin-MSTP)}} \\
        \midrule
        \multirow{2}{*}{\textbf{Method}} & \multicolumn{5}{c}{\textbf{Annotation Budget in Frames}} & \multirow{2}{*}{\textbf{nAULC}} & \multirow{2}{*}{\textbf{$p$-value}} \\
        \cmidrule{2-6}
        & \textbf{300k} & \textbf{400k} & \textbf{500k} & \textbf{600k} & \textbf{700k} & & \\
        \midrule
        Random & $77.4_{\pm0.3}$ & $71.9_{\pm0.7}$ & $65.7_{\pm1.5}$ & $59.4_{\pm1.3}$ & $54.6_{\pm0.0}$ & $68.8_{\pm0.8}$ & $0.025$ \\
        \midrule
        Entropy & $74.7_{\pm0.7}$ & $65.4_{\pm1.6}$ & $58.8_{\pm0.7}$ & $\mathbf{52.2_{\pm1.2}}$ & $48.2_{\pm0.9}$ & $63.5_{\pm0.7}$ & $0.703$ \\
        \midrule
        UNCAST & $79.5_{\pm0.9}$ & $78.4_{\pm6.1}$ & $68.6_{\pm1.4}$ & $63.8_{\pm1.0}$ & $58.2_{\pm0.3}$ & $72.3_{\pm0.9}$ & $0.008$ \\
        \midrule
        Core-set & $77.5_{\pm0.4}$ & $71.9_{\pm1.7}$ & $66.8_{\pm1.2}$ & $61.7_{\pm0.5}$ & $57.1_{\pm0.7}$ & $69.7_{\pm0.5}$ & $0.019$ \\
        \midrule
        Soft-Attn Core-set & $75.0_{\pm1.0}$ & $\mathbf{65.2_{\pm0.8}}$ & $\mathbf{58.3_{\pm0.8}}$ & $53.5_{\pm0.5}$ & $48.9_{\pm0.6}$ & $63.7_{\pm0.3}$ & $0.470$ \\
        \midrule
        \textbf{RAIDAL (Ours)} & $\mathbf{73.6_{\pm1.2}}$ & $66.1_{\pm0.8}$ & $58.4_{\pm1.0}$ & $52.6_{\pm0.7}$ & $\mathbf{48.0_{\pm0.7}}$ & $\mathbf{63.4_{\pm0.8}}$ & -- \\
        \bottomrule
    \end{tabular}
    }
    \caption{WER (\%) and nAULC on Isharah using Swin-MSTP. Results are mean $\pm$ std over 3 runs; $p$-values are obtained from $t$-tests against RAIDAL. The initial 200k-frame budget is omitted because it yields a shared baseline performance of $84.31\% \pm 0.37$.}
    \label{tab:wer_isharah_swin}
\end{table}

\subsection{Ablation Studies}

We conduct ablations on PHOENIX-2014 with CorrNet to isolate the contribution of RAIDAL's main design choices: decoder-based filtering, representation-based scoring, the number of decoded hypotheses used to construct the $\mathcal{TROI}$, and the Temporal Expansion Radius.

\noindent \textbf{Impact of Decoder-Based Filtering.} One of RAIDAL's contributions is its use of the CTC decoder to provide temporal structure to the acquisition function. Here we remove the CTC decoder as a filter and allow all timesteps to contribute to the video's acquisition score.

Table \ref{tab:ablations} (a) highlights a substantial performance degradation after removing this component, with the filterless variant obtaining worse WER than RAIDAL and Random at every acquisition cycle. Without the decoder's support, the acquisition score is again computed over all timesteps, allowing timesteps outside the decoder-supported regions to influence sample selection. This can lead to the selection of videos that may be distant in feature space for reasons unrelated to the evidence the decoder supports, reducing the quality of the labeled pool and validating RAIDAL's filter as a necessary part of the method.

\noindent \textbf{Applying Decoder-Based Filtering to Entropy.} To determine whether RAIDAL's performance derives mostly from its decoder-based filtering or from its representation-based scoring, we evaluate \textbf{RAIDAL-E}, a variant of Entropy that incorporates RAIDAL's filtering mechanism to restrict uncertainty estimation to the $\mathcal{TROI}$.

Table~\ref{tab:ablations} (b) shows that applying RAIDAL's filter to Entropy fails to help and instead makes it perform worse than both regular Entropy and Random. Together with the filtering ablation above, where removing RAIDAL's filter substantially degraded performance, RAIDAL-E's results show that the filter on its own does not automatically improve every acquisition strategy. While the filter restricts \emph{where} the acquisition function looks, the scoring function applied within that region is what ultimately determines whether the use of alignment peaks translates into a performance gain. Therefore, RAIDAL's performance does not stem from the filter alone, but rather from the \emph{combination of the filter and the underlying representation-based acquisition function}, both of which are necessary.

Additionally, the vocabulary analysis for RAIDAL-E shows that RAIDAL-E underperforms Entropy, RAIDAL, and Random Sampling across all budgets, despite covering a broader vocabulary than all three at every step (Table \ref{tab:supp_p14_corrnet}). This result further reinforces our earlier observation that vocabulary coverage alone may not be a reliable proxy for sample utility.

\noindent \textbf{Sensitivity to the Number of Decoded Hypotheses.} RAIDAL constructs the $\mathcal{TROI}$ from alignment peaks extracted from the top-$K$ hypotheses returned by the CTC decoder. The standard configuration uses $K=10$, matching the backbones' default beam search width of 10. To evaluate sensitivity to this choice, we test $K\in\{1,2,5\}$ on PHOENIX-2014/CorrNet while keeping the beam search width fixed at 10. Changing $K$ therefore does not modify the decoder, but only how many of the returned hypotheses contribute alignment peaks to the $\mathcal{TROI}$, with $K=1$ using only the highest-scoring hypothesis and larger values of $K$ including progressively more of the top decoded hypotheses.

Table \ref{tab:ablations} (c) shows that varying $K$ has little effect on performance, with nAULC values of 43.0, 43.1, and 43.3 for $K=1$, $K=2$, and $K=5$, compared with 43.3 for the default $K=10$. None of these differences are statistically significant ($p\geq0.252$), suggesting that RAIDAL is relatively insensitive to the number of decoded hypotheses used within the evaluated range.

\noindent \textbf{Sensitivity to the Temporal Expansion Radius ($R$).} To measure RAIDAL's sensitivity to the amount of temporal context retained around each alignment peak, we vary the canonical radius $R$. Setting $R=0$ removes temporal expansion entirely, while $R+1$, $R+2$, and $R+3$ progressively increase the retained context beyond the automatically derived value.

Table~\ref{tab:ablations} (d) shows that increasing $R$ ($R+1$, $R+2$, $R+3$) progressively degrades performance, suggesting that overly large windows reintroduce less useful timesteps into the scoring function. Although $R=0$ performs comparably to canonical RAIDAL on PHOENIX-2014, it can also behave less stably at the earlier Isharah budgets, particularly at 400k and 500k frames (Table \ref{tab:radius_isharah}). These results indicate that, while temporal expansion is not strictly required for performance, it can still provide additional local context and stability in the Isharah setting. We therefore keep $R$ as a stability parameter rather than as a performance component.

\begin{table}[H]
    \centering
    \small
    \setlength{\tabcolsep}{3pt}
    \resizebox{\linewidth}{!}{
    \begin{tabular}{lccccccccccc}
        \toprule
        \multirow{2}{*}{\textbf{Method}} & \multicolumn{9}{c}{\textbf{Annotation Budget in Frames}} & \multirow{2}{*}{\textbf{nAULC}} & \multirow{2}{*}{\textbf{$p$-value}} \\
        \cmidrule{2-10}
        & \textbf{30k} & \textbf{40k} & \textbf{50k} & \textbf{60k} & \textbf{80k} & \textbf{100k} & \textbf{120k} & \textbf{160k} & \textbf{200k} & & \\
        \midrule

        \multicolumn{12}{l}{\textit{Reference}} \\
        Random
        & $76.3_{\pm0.9}$ & $65.7_{\pm0.4}$ & $59.4_{\pm1.6}$ & $53.5_{\pm1.2}$ & $47.2_{\pm0.7}$ & $42.3_{\pm0.6}$ & $39.5_{\pm0.4}$ & $34.3_{\pm0.4}$ & $30.9_{\pm0.3}$ & $45.5_{\pm0.6}$ & $0.031$ \\
        Entropy & $76.1_{\pm0.4}$ & $66.2_{\pm0.4}$ & $58.2_{\pm1.2}$ & $55.9_{\pm1.8}$ & $47.6_{\pm0.9}$ & $42.3_{\pm0.2}$ & $39.7_{\pm0.5}$ & $34.4_{\pm0.6}$ & $30.4_{\pm0.3}$ & $45.8_{\pm0.2}$ & $<0.001$ \\
        \textbf{RAIDAL (Ours)}
        & $71.8_{\pm0.5}$ & $61.2_{\pm1.6}$ & $54.9_{\pm1.0}$ & $51.3_{\pm0.6}$ & $44.3_{\pm0.2}$ & $40.5_{\pm0.4}$ & $37.2_{\pm0.4}$ & $33.1_{\pm0.2}$ & $29.8_{\pm0.6}$ & $43.3_{\pm0.2}$ & -- \\
        \midrule

        \multicolumn{12}{l}{\textit{(a) Impact of Decoder-Based Filtering}} \\
        RAIDAL (No Filter)
        & $76.6_{\pm1.4}$ & $67.7_{\pm1.1}$ & $61.3_{\pm0.6}$ & $56.5_{\pm0.7}$ & $49.2_{\pm0.3}$ & $44.2_{\pm0.6}$ & $40.7_{\pm0.2}$ & $35.6_{\pm0.5}$ & $31.3_{\pm0.5}$ & $47.0_{\pm0.1}$ & $0.001$ \\
        \midrule

        \multicolumn{12}{l}{\textit{(b) Applying RAIDAL's filter to Entropy}} \\
        RAIDAL-E
        & $77.9_{\pm0.9}$ & $69.4_{\pm1.5}$ & $61.5_{\pm1.1}$ & $56.7_{\pm1.0}$ & $48.6_{\pm0.7}$ & $44.0_{\pm0.6}$ & $40.1_{\pm0.2}$ & $34.9_{\pm0.6}$ & $31.0_{\pm0.3}$ & $46.8_{\pm0.5}$ & $0.006$ \\
        \midrule

        \multicolumn{12}{l}{\textit{(c) Number of Decoded Hypotheses Used}} \\
        RAIDAL ($K=1$)
        & $71.6_{\pm1.2}$ & $61.4_{\pm0.8}$ & $55.3_{\pm0.9}$ & $51.0_{\pm0.5}$ & $44.5_{\pm1.0}$ & $39.5_{\pm0.9}$ & $36.6_{\pm0.2}$ & $32.3_{\pm0.1}$ & $30.0_{\pm0.3}$ & $43.0_{\pm0.3}$ & $0.252$ \\
        RAIDAL ($K=2$)
        & $70.3_{\pm1.3}$ & $61.8_{\pm1.3}$ & $55.4_{\pm0.9}$ & $51.0_{\pm0.5}$ & $44.3_{\pm0.3}$ & $40.2_{\pm0.6}$ & $36.9_{\pm0.4}$ & $32.5_{\pm0.2}$ & $29.9_{\pm0.2}$ & $43.1_{\pm0.3}$ & $0.307$ \\
        RAIDAL ($K=5$)
        & $71.0_{\pm0.5}$ & $60.9_{\pm0.2}$ & $54.5_{\pm0.3}$ & $51.0_{\pm0.2}$ & $44.9_{\pm0.5}$ & $40.7_{\pm0.2}$ & $37.2_{\pm0.4}$ & $32.9_{\pm0.3}$ & $30.1_{\pm0.3}$ & $43.3_{\pm0.1}$ & $0.977$ \\
        \midrule

        \multicolumn{12}{l}{\textit{(d) Temporal Expansion Radius}} \\
        RAIDAL ($R=0$)
        & $71.1_{\pm0.4}$ & $61.8_{\pm0.6}$ & $56.4_{\pm0.7}$ & $51.4_{\pm0.3}$ & $44.4_{\pm0.3}$ & $40.7_{\pm0.8}$ & $37.3_{\pm0.7}$ & $33.5_{\pm0.2}$ & $30.3_{\pm0.2}$ & $43.4_{\pm0.3}$ & $0.866$ \\
        RAIDAL ($R+1$)
        & $71.5_{\pm1.5}$ & $62.9_{\pm0.7}$ & $55.8_{\pm0.2}$ & $52.5_{\pm0.4}$ & $45.9_{\pm0.3}$ & $41.3_{\pm0.8}$ & $38.2_{\pm0.8}$ & $33.4_{\pm0.1}$ & $30.4_{\pm0.3}$ & $44.1_{\pm0.1}$ & $0.031$ \\
        RAIDAL ($R+2$)
        & $71.9_{\pm1.2}$ & $62.3_{\pm0.7}$ & $56.4_{\pm0.2}$ & $51.9_{\pm0.3}$ & $46.6_{\pm0.5}$ & $41.9_{\pm0.3}$ & $39.0_{\pm0.3}$ & $34.2_{\pm0.6}$ & $30.6_{\pm0.5}$ & $44.6_{\pm0.2}$ & $0.044$ \\
        RAIDAL ($R+3$)
        & $70.6_{\pm0.3}$ & $64.1_{\pm0.3}$ & $58.3_{\pm0.7}$ & $53.0_{\pm0.8}$ & $47.3_{\pm0.5}$ & $42.9_{\pm0.6}$ & $39.2_{\pm0.2}$ & $34.1_{\pm0.2}$ & $30.6_{\pm0.3}$ & $45.0_{\pm0.2}$ & $0.006$ \\
        \bottomrule
    \end{tabular}
    }
    \caption{Ablation studies on PHOENIX-2014 using CorrNet. Results are mean WER $\pm$ std over 3 runs. The $p$-values are obtained from $t$-tests against the default RAIDAL configuration with $K=10$ and the canonical radius. (a) removes decoder-based filtering and allows all timesteps to contribute to the acquisition score. (b) applies RAIDAL's filter to Entropy. (c) varies the number of decoded hypotheses used to construct the $\mathcal{TROI}$. (d) varies the Temporal Expansion Radius $R$.}
    \label{tab:ablations}
\end{table}

\begin{table}[H]
    \centering
    \small
    \resizebox{\linewidth}{!}{
    \begin{tabular}{lccccccc}
        \toprule
        \multicolumn{8}{c}{\textbf{Isharah: Temporal Expansion Radius Ablation}} \\
        \midrule
        \multirow{2}{*}{\textbf{Method}} & \multicolumn{5}{c}{\textbf{Annotation Budget in Frames}} & \multirow{2}{*}{\textbf{nAULC}} & \multirow{2}{*}{\textbf{$p$-value}} \\
        \cmidrule{2-6}
        & \textbf{300k} & \textbf{400k} & \textbf{500k} & \textbf{600k} & \textbf{700k} & & \\
        \midrule
        RAIDAL ($R=0$) & $74.2_{\pm0.8}$ & $67.5_{\pm2.2}$ & $60.1_{\pm2.9}$ & $52.7_{\pm0.5}$ & $\mathbf{47.8_{\pm1.2}}$ & $64.1_{\pm0.8}$ & $0.111$ \\
        \textbf{RAIDAL (Canonical $R$)} & $\mathbf{73.6_{\pm1.2}}$ & $\mathbf{66.1_{\pm0.8}}$ & $\mathbf{58.4_{\pm1.0}}$ & $\mathbf{52.6_{\pm0.7}}$ & $48.0_{\pm0.7}$ & $\mathbf{63.4_{\pm0.8}}$ & -- \\
        \bottomrule
    \end{tabular}
    }
    \caption{Temporal Expansion Radius ablation on Isharah using Swin-MSTP. Results are mean $\pm$ std over 3 runs; the $p$-value is obtained from a paired $t$-test against canonical RAIDAL. The initial 200k-frame budget is omitted because it yields a shared baseline performance of $84.31\% \pm 0.37$.}
    \label{tab:radius_isharah}
\end{table}

\subsection{Additional Results and Analyses}
The supplementary material provides further analyses of computational cost, low-budget robustness, selection bias, practical annotation savings, and the effect of applying RAIDAL's filter to CTC-BADGE. RAIDAL remains effective when initialized with a substantially smaller frame budget and has shorter acquisition time and lower VRAM usage than Soft-Attention Core-set. Applying RAIDAL's filtering to CTC-BADGE also improves nAULC numerically, suggesting that decoder-based filtering may be useful beyond representation-based acquisition functions, although this difference is not statistically significant.

% The supplementary material provides additional results and analyses of computational cost, low-budget robustness, selection bias, and practical annotation savings. RAIDAL remains effective when initialized with a substantially smaller frame budget and has shorter acquisition time and lower VRAM usage than Soft-Attention Core-set. RAIDAL's WER improvements on PHOENIX-T also reduce annotation time by an estimated 9.2 to 18.4 hours.

\section{Conclusion}

We presented, to the best of our knowledge, the first study of active learning for continuous sign language recognition. The central challenge addressed in this work arises from the weak temporal alignment of CSLR videos, where sign executions are interleaved with rest poses, sign transitions, and temporally redundant timesteps. Since supervision is only available as an ordered gloss sequence, an acquisition function cannot safely assume that every timestep is equally informative for sample selection.

Building on these observations, we introduced RAIDAL, a redundancy-aware active learning strategy that repurposes CTC decoder alignment peaks as a source of temporal structure. RAIDAL restricts feature-space diversity scoring to the temporal regions of interest these peaks identify, directing the acquisition function to focus on the temporal evidence the decoder supports rather than on the full unfiltered sequence. Across PHOENIX-2014, PHOENIX-T, and Isharah-1000, using CNN-based and Transformer-based models, RAIDAL obtains its strongest gains over competing baselines in large-vocabulary, budget-limited settings while remaining competitive when the vocabulary is smaller and annotation budgets are larger.

More broadly, our results suggest that active learning acquisition functions do not need to allow the full sequence to influence sample selection indiscriminately. Instead, focusing selection on the segments the model itself identifies as supporting its predictions can provide a more reliable basis for estimating a sample's informativeness.

\subsection{Future Work}
\label{subsec:futurework}

Several directions remain open for future research. First, the PHOENIX datasets and Isharah differ simultaneously in vocabulary size and visual variation, making it difficult to isolate which of these factors explains the narrower performance gap observed on Isharah. Evaluating RAIDAL on a CSLR dataset that combines a large vocabulary with unconstrained visual conditions would help isolate the role of vocabulary size in RAIDAL's performance. Second, our experiments are limited to the CorrNet and Swin-MSTP backbones, and evaluating RAIDAL on more architectures would test whether the method transfers to different feature spaces. Finally, RAIDAL is currently limited to CTC-based CSLR because the $\mathcal{TROI}$ is constructed from CTC alignment peaks, leaving its extension to non-CTC models and alternative alignment mechanisms as an open direction.

RAIDAL's acquisition function also offers an opportunity for further work. RAIDAL, RAIDAL-E, and applying the filter to CTC-BADGE shows that changing the scoring function while keeping the filtering strategy fixed produces substantially different levels of improvement, but so far this has only been tested with two alternative scoring functions (Entropy and CTC-BADGE). Evaluating RAIDAL's filter with a broader range of acquisition functions could reveal how alignment peaks affect each one, and whether other acquisition functions also benefit from the filter in the same way representation-based scoring does. Beyond CSLR, RAIDAL could also be evaluated in other CTC-based recognition tasks such as long-form automatic speech recognition or visual speech recognition, where activations can be ambiguous and alignment peaks may provide a useful signal for active learning.

\section*{Acknowledgements}

We gratefully acknowledge support from CAPES, CNPq, and FAPEMIG.

% ============================================================
% References
% ============================================================
\clearpage
\bibliography{egbib}

% ============================================================
% Supplementary Material
% ============================================================
\clearpage
\appendix

\section*{Supplementary Material}

This supplementary material provides additional experimental analyses and vocabulary statistics supporting the results presented in the main paper.

\section{Computational Cost}

Because the timestep-level nearest-neighbor scoring used by RAIDAL requires comparing features from the unlabeled and labeled pools, its computational cost can grow quickly with the number of retained features. We therefore measure the acquisition cost of RAIDAL against Core-set, which collapses the temporal dimension into a single global feature vector per video, and against Soft-Attention Core-set, which retains all timestep-level features. We used a single RTX 4090 to evaluate the first PHOENIX-2014/CorrNet acquisition cycle, where the unlabeled pool is largest. This setting contains 139 labeled videos (20k frames / 5,078 labeled features), 5,532 unlabeled videos (780k frames / 196,978 unlabeled features), and a 10k-frame acquisition budget. We report VRAM usage and acquisition time both with early stopping, where selection stops once the annotation budget is reached, and without early stopping, where the complete candidate pool is scored. All methods use exact nearest-neighbor search without feature subsampling or approximate nearest neighbors.

Table \ref{tab:acquisition_cost} shows that RAIDAL requires 58 seconds and 750 MB of VRAM with early stopping, compared with 96 seconds and 1250 MB for Soft-Attention Core-set. Although RAIDAL requires 26 seconds to CTC-decode the entire unlabeled pool in order to obtain its alignment peaks, its $\mathcal{TROI}$ filter removes timesteps outside the decoder-aligned regions before nearest-neighbor scoring, reducing the number of feature comparisons and the memory usage relative to evaluating every timestep. This difference becomes larger when scoring the full pool, with RAIDAL requiring 20 minutes, compared with 62 minutes for Soft-Attention Core-set. Core-set remains substantially cheaper (3 seconds with early stopping and 41 seconds on the full pool) because it operates on a single global feature vector per video.

\begin{table}[H]
    \centering
    \small
    \setlength{\tabcolsep}{3pt}
    \resizebox{\linewidth}{!}{
    \begin{tabular}{l|cc|c}
        \toprule
        \multicolumn{4}{c}{\textbf{PHOENIX-2014 (CorrNet)}} \\
        \midrule
        \multirow{2}{*}{\textbf{Method}} & \multicolumn{2}{c|}{\textbf{Acquisition Time}} & \multirow{2}{*}{\textbf{VRAM}} \\
        \cmidrule{2-3}
         & \textbf{With early stopping} & \textbf{Without early stopping} & \\
        \midrule
        Core-set & $3$ seconds & $41$ seconds & $520$ MB \\
        \midrule
        Soft-Attn Core-set & $96$ seconds & $62$ minutes & $1250$ MB\\
        \midrule
        \textbf{RAIDAL (Ours)} & $58$ seconds ($26$ s decode + $32$ s selection) & $20$ minutes (including $26$ s decode) & $750$ MB \\
        \bottomrule
    \end{tabular}
    }
    \caption[Computational cost of acquisition]
    {Computational cost of the acquisition step on PHOENIX-2014 using CorrNet and a single RTX 4090.}
    \label{tab:acquisition_cost}
\end{table}

\section{Robustness in the Low-Budget Setting}
\label{sec:robustness}

Because RAIDAL relies on CTC decoder alignment peaks, we additionally evaluate whether its filtering remains useful when the recognition model is weaker. We therefore repeat the PHOENIX-2014/CorrNet experiment under a substantially more restrictive initialization, reducing the initial labeled pool from 20k to 5k frames and querying in 5k-frame increments up to 25k frames (new budgets: 5k$\to$10k$\to$15k$\to$20k$\to$25k; $n$=3 replications; the full training set has 8.9 hours of footage (800k frames); 5k frames = 3.3 minutes). This initial pool contains 401 total gloss tokens covering 155 classes, of which 82 (53\%) occur only once. The initial model has a WER of 90.52\%.

Table \ref{tab:wer_phoenix14_lowbudget} shows that, despite this weak initialization, RAIDAL achieves an nAULC of 81.7, with statistically significant gains over both Random Sampling ($p=0.010$) and its own no-filter variant from ablation (a) ($p=0.005$). The advantage is already visible from the first acquisition cycle, where RAIDAL obtains 85.6\% WER compared with 88.6\% for Random Sampling and 88.3\% for the no-filter variant. These results suggest that alignment peaks can provide a useful signal even when the initial model is very weak, as removing decoder-based filtering consistently reduces performance relative to RAIDAL.

\begin{table}[H]
    \centering
    \small
    \setlength{\tabcolsep}{3pt}
    \resizebox{\linewidth}{!}{
    \begin{tabular}{l|cccc|cc}
        \toprule
        \multicolumn{7}{c}{\textbf{PHOENIX-2014 (CorrNet, Low-Budget Regime)}} \\
        \midrule
        \multirow{2}{*}{\textbf{Method}} & \multicolumn{4}{c|}{\textbf{Annotation Budget in Frames}} & \multirow{2}{*}{\textbf{nAULC}} & \multirow{2}{*}{\textbf{$p$-value}} \\
        \cmidrule{2-5}
         & \textbf{10k} & \textbf{15k} & \textbf{20k} & \textbf{25k} & & \\
        \midrule
        Random & $88.6_{\pm0.4}$ & $85.8_{\pm0.5}$ & $82.3_{\pm0.9}$ & $80.1_{\pm0.3}$ & $85.5_{\pm0.2}$ & $0.010$ \\
        \midrule
        RAIDAL (No Filter) & $88.3_{\pm0.5}$ & $86.3_{\pm0.6}$ & $83.9_{\pm0.7}$ & $80.2_{\pm0.9}$ & $86.0_{\pm0.1}$ & $0.005$ \\
        \midrule
        \textbf{RAIDAL (Ours)} & $\mathbf{85.6_{\pm0.6}}$ & $\mathbf{81.3_{\pm0.4}}$ & $\mathbf{77.7_{\pm0.8}}$ & $\mathbf{74.1_{\pm0.9}}$ & $\mathbf{81.7_{\pm0.5}}$ & -- \\
        \bottomrule
    \end{tabular}
    }
    \caption[Low-budget robustness on PHOENIX-2014]
    {WER (\%) and nAULC on PHOENIX-2014 using CorrNet under a low-budget initialization. Results are mean $\pm$ std over 3 runs; $p$-values are obtained from $t$-tests against RAIDAL. The initial 5k-frame budget is omitted because it yields a shared baseline performance of 90.52\% $\pm$ 0.46.}
    \label{tab:wer_phoenix14_lowbudget}
\end{table}

\section{Selection Bias Across Acquisition Cycles}
\label{sec:selection_bias}

Since RAIDAL restricts the acquisition function to decoder-supported regions, we evaluate whether these regions introduce a preference toward frequent glosses, particularly during the early acquisition cycles when the model is still weak. We analyze this behavior on PHOENIX-2014 with CorrNet by measuring how frequently head glosses are selected relative to their availability in the candidate pool. We define the head vocabulary as the top 5\% most frequent glosses in PHOENIX-2014, and measure Head-Token Selection Bias as the percentage of head tokens in the selected batch minus their percentage in the candidate pool immediately before selection. Therefore, positive values indicate a stronger selection bias toward frequent glosses relative to their availability, while negative values indicate a bias away from frequent glosses.

Table \ref{tab:selection_bias} shows that RAIDAL initially favors selecting frequent glosses, reaching biases of $+3.5$, $+4.7$, and $+3.7$ percentage points during the first three cycles, but this preference disappears by the fourth cycle and becomes negative later in acquisition, indicating a gradual shift away from frequent glosses as the labeled pool grows. More interestingly, RAIDAL-E exhibits the opposite behavior from the start despite using the same filtering logic as RAIDAL. RAIDAL-E consistently favors less frequent glosses, behaving similarly to Entropy and UNCAST. This further supports our observation that decoder-based filtering does not enforce a particular selection pattern, since keeping the filtering mechanism unchanged while replacing the scoring function leads to substantially different acquisition profiles.

We also highlight that RAIDAL's slight early preference toward frequent glosses suggests that Head-Token Selection Bias may not be a reliable proxy for sample utility. RAIDAL already achieves the best WER after the first acquisition cycle despite showing a slight bias toward frequent glosses, while methods with both similar and opposite frequency profiles perform worse (Table \ref{tab:wer_phoenix14}). Taken together with the vocabulary analysis, these results suggest that performance is more closely associated with maintaining sufficient repeated exposure to selected glosses than with the direction of the frequency bias itself.

\begin{table}[H]
    \centering
    \small
    \setlength{\tabcolsep}{3pt}
    \resizebox{\linewidth}{!}{
    \begin{tabular}{lccccccccc}
        \toprule
        \multicolumn{10}{c}{\textbf{PHOENIX-2014 (CorrNet)}} \\
        \midrule
        \multirow{2}{*}{\textbf{Method}} & 
        \multicolumn{9}{c}{\textbf{Annotation Budget in Frames}} \\
        \cmidrule{2-10} & \textbf{30k}  & \textbf{40k}  & \textbf{50k}  & \textbf{60k}  & \textbf{80k}  & \textbf{100k}  & \textbf{120k}  & \textbf{160k} & \textbf{200k} \\
        \midrule
        Random
        & $+1.7_{\pm1.8}$ & $+0.2_{\pm2.0}$ & $+0.7_{\pm0.6}$ & $-0.5_{\pm4.0}$ & $+1.3_{\pm1.6}$ & $-0.5_{\pm2.4}$ & $-0.1_{\pm1.7}$ & $+0.1_{\pm0.5}$ & $+0.1_{\pm0.9}$ \\
        \midrule
        Entropy
        & $-3.9_{\pm3.0}$ & $-5.4_{\pm2.2}$ & $-10.8_{\pm0.6}$ & $-10.1_{\pm1.6}$ & $-9.5_{\pm4.1}$ & $-11.0_{\pm3.3}$ & $-12.0_{\pm0.5}$ & $-8.5_{\pm2.6}$ & $-8.6_{\pm1.4}$ \\
        \midrule
        UNCAST
        & $-7.8_{\pm0.2}$ & $-12.3_{\pm3.1}$ & $-9.2_{\pm1.4}$ & $-14.4_{\pm2.3}$ & $-13.7_{\pm2.1}$ & $-15.7_{\pm0.4}$ & $-10.7_{\pm1.3}$ & $-11.2_{\pm0.9}$ & $-8.9_{\pm0.5}$ \\
        \midrule
        Core-set
        & $-2.6_{\pm1.2}$ & $-3.5_{\pm2.7}$ & $-5.1_{\pm1.1}$ & $-6.0_{\pm3.6}$ & $-7.2_{\pm2.3}$ & $-6.7_{\pm0.5}$ & $-7.1_{\pm2.0}$ & $-5.2_{\pm1.3}$ & $-5.1_{\pm0.7}$ \\
        \midrule
        Soft-Attn Core-set
        & $+3.2_{\pm0.9}$ & $+2.0_{\pm1.8}$ & $+1.1_{\pm1.2}$ & $+0.1_{\pm1.2}$ & $-0.4_{\pm1.9}$ & $-0.9_{\pm1.4}$ & $-1.5_{\pm2.2}$ & $-1.4_{\pm1.1}$ & $-3.2_{\pm0.9}$ \\
        \midrule
        CTC-BADGE
        & $-0.4_{\pm2.4}$ & $-2.4_{\pm3.0}$ & $-1.1_{\pm2.5}$ & $+0.6_{\pm0.1}$ & $-1.2_{\pm1.4}$ & $-0.9_{\pm1.7}$ & $-0.3_{\pm0.4}$ & $-1.5_{\pm0.3}$ & $-3.4_{\pm1.3}$ \\
        \midrule
        \textbf{RAIDAL}
        & $+3.5_{\pm1.7}$ & $+4.7_{\pm0.1}$ & $+3.7_{\pm0.4}$ & $-0.7_{\pm1.8}$ & $-0.2_{\pm0.9}$ & $-3.8_{\pm0.6}$ & $-5.7_{\pm2.2}$ & $-3.8_{\pm1.0}$ & $-5.1_{\pm1.6}$ \\
        \midrule
        \textbf{RAIDAL-E}
        & $-7.7_{\pm2.4}$ & $-9.3_{\pm3.0}$ & $-13.2_{\pm0.8}$ & $-11.0_{\pm0.9}$ & $-12.4_{\pm0.9}$ & $-14.2_{\pm1.1}$ & $-11.5_{\pm0.9}$ & $-11.2_{\pm1.0}$ & $-8.7_{\pm1.3}$ \\
        \bottomrule
    \end{tabular}
    }
    \caption[Selection bias across acquisition cycles]
    {Head-Token Selection Bias analysis on PHOENIX-2014 using CorrNet across acquisition cycles. Results are reported in percentage points as mean $\pm$ std over 3 runs.}
    \label{tab:selection_bias}
\end{table}

\section{Practical Impact: Translating Savings to Annotation Hours}

Beyond WER reductions, the main goal of active learning is to reduce human annotation effort. Duarte \etal report that gloss annotation for How2Sign required, on average, one hour of expert work to annotate 90 seconds of video \cite{annotationcost3}, corresponding to approximately 40 minutes of annotation work per minute of footage. Stein \etal report that an annotator with two months of experience took, on average, 4 hours to annotate three RWTH-Phoenix weather forecast videos of roughly one minute each \cite{annotationcost4}, corresponding to approximately 80 minutes of annotation work per minute of footage. 

We ground our estimate in these two measurements because they are the most directly relevant to our setting. Duarte \etal's figure comes from How2Sign, a continuous sign language dataset, and Stein \etal's figure comes from RWTH-Phoenix, which is from the same family as the PHOENIX datasets used in this work. The near twofold spread between the two figures (40 vs. 80 minutes per minute of footage) also motivates us to treat them as an interval rather than as a single point estimate. We therefore adopt this 40--80-minute range as an estimate of the amount of time that it takes to annotate one minute of CSLR footage.

We quantify RAIDAL's practical impact on CorrNet/PHOENIX-T against Random Sampling, which was RAIDAL's closest-performing competitor in this setting. In PHOENIX-T, videos are recorded at 25 FPS and one minute of footage corresponds to 1,500 frames. As shown in Table~\ref{tab:practical_impact}, RAIDAL reaches both the $\le 45\%$ and $\le 40\%$ WER targets one annotation cycle earlier than Random Sampling. Since at these budgets one cycle corresponds to 20,700 frames, this represents approximately 13.8 minutes of footage that would not need to be annotated, reducing annotation labor by an estimated 9.2 to 18.4 hours. At the $\le 30\%$ milestone, RAIDAL reaches the target within the evaluated budget, while Random Sampling does not.

\begin{table}[H]
    \centering
    \small
    \resizebox{\linewidth}{!}{
    \begin{tabular}{llcc|c}
        \toprule
        \textbf{Target WER} & \textbf{Method} & \textbf{Frames} & \textbf{Video Min.} & \textbf{Labor Avoided} \\
        \midrule
        \multirow{2}{*}{$\le 45\%$} 
        & Random Sampling & 103k & 69 & \multirow{2}{*}{Between 9.2 and 18.4 hours} \\
        & \textbf{RAIDAL (Ours)} & \textbf{82k}  & \textbf{55} & \\
        \midrule
        \multirow{2}{*}{$\le 40\%$} 
        & Random Sampling & 124k & 83 & \multirow{2}{*}{Between 9.2 and 18.4 hours} \\
        & \textbf{RAIDAL (Ours)} & \textbf{103k} & \textbf{69} & \\
        \midrule
        \multirow{2}{*}{$\le 30\%$} 
        & Random Sampling & $30.7\%$ at 207k & $> 138$ & \multirow{2}{*}{RAIDAL reaches target and Random does not.} \\
        & \textbf{RAIDAL (Ours)} & \textbf{207k} & \textbf{138} & \\
        \bottomrule
    \end{tabular}
    }
    \caption{Practical impact of RAIDAL on CorrNet/PHOENIX-T across WER milestones.}
    \label{tab:practical_impact}
\end{table}

% INCLUIR ESSA ANÁLISE APENAS NA VERSÃO DO ARXIV !! 

\section{Applying RAIDAL's Filtering to CTC-BADGE}

We additionally evaluate whether RAIDAL's decoder-based filtering can support a gradient-based acquisition function. Since CTC-BADGE constructs its gradient embedding using contributions from the full sequence, timesteps outside decoder-supported regions can still influence sample selection. We therefore apply the same filtering used by RAIDAL to CTC-BADGE and restrict the gradient embedding to timestep contributions inside the $\mathcal{TROI}$. The CTC loss is still computed over the full sequence, while only the retained timesteps contribute to the gradient embedding before k-means++ selection.
 
Applying decoder-based filtering to CTC-BADGE improves its nAULC from $45.2\pm0.4$ to $44.5\pm0.1$ (Table \ref{tab:ctc_badge_filtering}), although this improvement is not statistically significant ($p=0.107$). RAIDAL still performs significantly better than the filtered CTC-BADGE variant ($43.3\pm0.2$ vs. $44.5\pm0.1$, $p=0.012$). These results indicate that decoder-based filtering may also be beneficial for acquisition functions outside the representation-based family, although filtering alone is not sufficient to produce a significant improvement in this setting. The significant difference between RAIDAL and filtered CTC-BADGE supports our previous observation that the acquisition function itself remains an important component in addition to the filtering strategy.

\begin{table}[H]
     \centering
     \small
     \setlength{\tabcolsep}{3pt}
     \resizebox{\linewidth}{!}{
     \begin{tabular}{lccccccccccc}
         \toprule
         \multicolumn{12}{c}{\textbf{PHOENIX-2014 (CorrNet)}} \\
         \midrule
         \multirow{2}{*}{\textbf{Method}} & \multicolumn{9}{c}{\textbf{Annotation Budget in Frames}} & \multirow{2}{*}{\textbf{nAULC}} & \multirow{2}{*}{\textbf{$p$-value}} \\
         \cmidrule{2-10}
         & \textbf{30k} & \textbf{40k} & \textbf{50k} & \textbf{60k} & \textbf{80k} & \textbf{100k} & \textbf{120k} & \textbf{160k} & \textbf{200k} & & \\
         \midrule
         CTC-BADGE
         & $74.7_{\pm0.6}$ & $65.2_{\pm1.3}$ & $57.8_{\pm0.9}$ & $53.8_{\pm0.9}$ & $47.6_{\pm0.5}$ & $42.4_{\pm0.1}$ & $38.7_{\pm0.7}$ & $34.0_{\pm0.3}$ & $30.9_{\pm0.4}$ & $45.2_{\pm0.4}$ & $0.037$ \\
         \midrule
         CTC-BADGE + $\mathcal{TROI}$
         & $74.4_{\pm1.1}$ & $63.1_{\pm0.5}$ & $57.2_{\pm0.0}$ & $52.5_{\pm0.4}$ & $46.2_{\pm0.2}$ & $41.1_{\pm0.4}$ & $38.1_{\pm0.4}$ & $33.8_{\pm0.2}$ & $30.9_{\pm0.3}$ & $44.5_{\pm0.1}$ & $0.012$ \\
         \midrule
         \textbf{RAIDAL (Ours)}
         & $\mathbf{71.8_{\pm0.5}}$ & $\mathbf{61.2_{\pm1.6}}$ & $\mathbf{54.9_{\pm1.0}}$ & $\mathbf{51.3_{\pm0.6}}$ & $\mathbf{44.3_{\pm0.2}}$ & $\mathbf{40.5_{\pm0.4}}$ & $\mathbf{37.2_{\pm0.4}}$ & $\mathbf{33.1_{\pm0.2}}$ & $\mathbf{29.8_{\pm0.6}}$ & $\mathbf{43.3_{\pm0.2}}$ & -- \\
         \bottomrule
     \end{tabular}
     }
     \caption{WER (\%) and nAULC on PHOENIX-2014 using CorrNet when applying RAIDAL's $\mathcal{TROI}$ to CTC-BADGE. Results are mean $\pm$ std over 3 runs. The $p$-values are obtained from $t$-tests against RAIDAL. The difference between CTC-BADGE and CTC-BADGE + $\mathcal{TROI}$ is not statistically significant ($p=0.107$).}
     \label{tab:ctc_badge_filtering}
 \end{table}

\section{Vocabulary Statistics}

This section presents vocabulary statistics across annotation budgets to further characterize the sampling behavior of each acquisition method. We report the number of unique glosses acquired, vocabulary coverage, singleton glosses, and mean gloss frequency. All values are means over three independent replications.

\newpage

\begin{table}[H]
    \centering
    \small
    \setlength{\tabcolsep}{5pt}
    \resizebox{\linewidth}{!}{
    \begin{tabular}{l|ccccccccc}
        \toprule
        \multirow{2}{*}{\textbf{Method}} & \multicolumn{9}{c}{\textbf{Annotation Budget in Frames}} \\
        \cmidrule{2-10}
        & \textbf{30k} & \textbf{40k} & \textbf{50k} & \textbf{60k} & \textbf{80k} & \textbf{100k} & \textbf{120k} & \textbf{160k} & \textbf{200k} \\
        \midrule
        \multicolumn{10}{l}{\textit{(a) Unique Glosses Acquired}} \\
        Random                      & 369 & 414 & 454 & 486 & 539 & 580 & 625 & 696 & 757 \\
        Entropy                     & 383 & 443 & 502 & 560 & 657 & 728 & 791 & 881 & 958 \\
        UNCAST                      & 383 & 455 & 501 & 551 & 643 & 727 & 793 & 889 & 969 \\
        Core-set                    & 380 & 436 & 487 & 529 & 610 & 677 & 720 & 812 & 880 \\
        Soft-Attn Core-set          & 360 & 400 & 436 & 465 & 530 & 584 & 631 & 712 & 793 \\
        CTC-BADGE                   & 370 & 415 & 456 & 493 & 549 & 601 & 647 & 728 & 801 \\
        \textbf{RAIDAL}             & 361 & 398 & 426 & 466 & 534 & 598 & 659 & 760 & 832 \\
        \midrule
        \textbf{RAIDAL-E}           & 387 & 455 & 526 & 572 & 662 & 743 & 799 & 900 & 976 \\
        \midrule
        \multicolumn{10}{l}{\textit{(b) Vocabulary Coverage (\%)}} \\
        Random                      & 29.9 & 33.6 & 36.9 & 39.5 & 43.8 & 47.1 & 50.8 & 56.6 & 61.5 \\
        Entropy                     & 31.1 & 36.0 & 40.8 & 45.5 & 53.3 & 59.1 & 64.2 & 71.6 & 77.8 \\
        UNCAST                      & 31.1 & 37.0 & 40.7 & 44.8 & 52.2 & 59.1 & 64.4 & 72.2 & 78.7 \\
        Core-set                    & 30.9 & 35.4 & 39.6 & 43.0 & 49.6 & 55.0 & 58.5 & 65.9 & 71.5 \\
        Soft-Attn Core-set          & 29.2 & 32.5 & 35.4 & 37.8 & 43.1 & 47.4 & 51.3 & 57.8 & 64.4 \\
        CTC-BADGE                   & 30.1 & 33.7 & 37.0 & 40.1 & 44.6 & 48.8 & 52.6 & 59.1 & 65.1 \\
        \textbf{RAIDAL}             & 29.3 & 32.3 & 34.6 & 37.9 & 43.4 & 48.6 & 53.6 & 61.8 & 67.6 \\
        \midrule
        \textbf{RAIDAL-E}           & 31.4 & 37.0 & 42.7 & 46.5 & 53.8 & 60.4 & 64.9 & 73.1 & 79.3 \\
        \midrule
        \multicolumn{10}{l}{\textit{(c) Singletons (glosses appearing once)}} \\
        Random                      & 149 & 158 & 172 & 179 & 191 & 203 & 212 & 241 & 257 \\
        Entropy                     & 151 & 170 & 188 & 204 & 235 & 252 & 276 & 309 & 327 \\
        UNCAST                      & 154 & 183 & 196 & 210 & 239 & 258 & 281 & 316 & 341 \\
        Core-set                    & 143 & 160 & 177 & 190 & 219 & 240 & 250 & 275 & 302 \\
        Soft-Attn Core-set          & 132 & 146 & 150 & 156 & 180 & 194 & 209 & 233 & 266 \\
        CTC-BADGE                   & 144 & 154 & 171 & 181 & 192 & 207 & 222 & 247 & 271 \\
        \textbf{RAIDAL}             & 135 & 142 & 145 & 163 & 191 & 206 & 228 & 265 & 284 \\
        \midrule
        \textbf{RAIDAL-E}           & 156 & 180 & 207 & 215 & 242 & 265 & 281 & 313 & 342 \\
        \midrule
        \multicolumn{10}{l}{\textit{(d) Mean Gloss Frequency (mean number of occurrences per acquired gloss)}} \\
        Random                      & 6.7 & 7.9 & 9.0 & 10.1 & 12.1 & 14.1 & 15.7 & 18.8 & 21.6 \\
        Entropy                     & 6.7 & 7.9 & 8.9 & 9.6 & 11.1 & 12.5 & 13.7 & 16.2 & 18.5 \\
        UNCAST                      & 6.4 & 7.1 & 7.9 & 8.5 & 9.8 & 10.8 & 11.9 & 14.2 & 16.3 \\
        Core-set                    & 6.7 & 7.8 & 8.7 & 9.6 & 11.0 & 12.4 & 13.9 & 16.3 & 18.8 \\
        Soft-Attn Core-set          & 7.4 & 9.1 & 10.6 & 12.1 & 14.4 & 16.4 & 18.2 & 21.4 & 23.9 \\
        CTC-BADGE                   & 6.7 & 8.0 & 9.1 & 10.1 & 12.1 & 13.8 & 15.3 & 18.2 & 20.7 \\
        \textbf{RAIDAL}             & 7.4 & 9.1 & 10.7 & 11.9 & 13.9 & 15.6 & 16.9 & 19.5 & 22.1 \\
        \midrule
        \textbf{RAIDAL-E}           & 6.4 & 7.2 & 7.9 & 8.8 & 10.2 & 11.4 & 12.7 & 15.0 & 17.2 \\
        \bottomrule
    \end{tabular}
    }
\caption{Vocabulary statistics for CorrNet on PHOENIX-2014 across annotation budgets. Values are means over three independent replications.}    
\label{tab:supp_p14_corrnet}
\end{table}
 
\begin{table}[H]
    \centering
    \small
    \setlength{\tabcolsep}{5pt}
    \resizebox{\linewidth}{!}{
    \begin{tabular}{l|ccccccccc}
        \toprule
        \multirow{2}{*}{\textbf{Method}} & \multicolumn{9}{c}{\textbf{Annotation Budget in Frames}} \\
        \cmidrule{2-10}
        & \textbf{31k} & \textbf{41k} & \textbf{52k} & \textbf{62k} & \textbf{82k} & \textbf{103k} & \textbf{124k} & \textbf{165k} & \textbf{207k} \\
        \midrule
        \multicolumn{10}{l}{\textit{(a) Unique Glosses Acquired}} \\
        Random                      & 329 & 371 & 406 & 435 & 490 & 528 & 560 & 625 & 682 \\
        Entropy                     & 345 & 395 & 455 & 501 & 584 & 673 & 742 & 827 & 889 \\
        UNCAST                      & 370 & 430 & 482 & 524 & 617 & 699 & 768 & 849 & 908 \\
        Core-set                    & 346 & 394 & 447 & 484 & 548 & 609 & 663 & 753 & 823 \\
        Soft-Attn Core-set          & 336 & 374 & 407 & 438 & 476 & 530 & 574 & 654 & 710 \\
        \textbf{RAIDAL}             & 332 & 371 & 407 & 442 & 502 & 562 & 615 & 696 & 764 \\
        \midrule
        \multicolumn{10}{l}{\textit{(b) Vocabulary Coverage (\%)}} \\
        Random                      & 30.3 & 34.2 & 37.4 & 40.1 & 45.1 & 48.7 & 51.6 & 57.6 & 62.8 \\
        Entropy                     & 31.8 & 36.4 & 41.9 & 46.2 & 53.8 & 62.0 & 68.4 & 76.2 & 82.0 \\
        UNCAST                      & 34.1 & 39.6 & 44.4 & 48.3 & 56.8 & 64.4 & 70.8 & 78.2 & 83.7 \\
        Core-set                    & 31.9 & 36.3 & 41.2 & 44.6 & 50.5 & 56.2 & 61.1 & 69.4 & 75.9 \\
        Soft-Attn Core-set          & 30.9 & 34.4 & 37.5 & 40.4 & 43.9 & 48.9 & 52.9 & 60.2 & 65.5 \\
        \textbf{RAIDAL}             & 30.6 & 34.2 & 37.5 & 40.7 & 46.2 & 51.8 & 56.7 & 64.1 & 70.4 \\
        \midrule
        \multicolumn{10}{l}{\textit{(c) Singletons (glosses appearing once)}} \\
        Random                      & 125 & 142 & 157 & 165 & 180 & 192 & 198 & 217 & 236 \\
        Entropy                     & 134 & 154 & 182 & 193 & 220 & 263 & 285 & 297 & 309 \\
        UNCAST                      & 151 & 178 & 196 & 209 & 248 & 269 & 296 & 316 & 326 \\
        Core-set                    & 135 & 152 & 175 & 189 & 211 & 233 & 248 & 277 & 300 \\
        Soft-Attn Core-set          & 134 & 145 & 155 & 166 & 169 & 186 & 199 & 230 & 247 \\
        \textbf{RAIDAL}             & 125 & 139 & 151 & 166 & 183 & 205 & 226 & 249 & 271 \\
        \midrule
        \multicolumn{10}{l}{\textit{(d) Mean Gloss Frequency (mean number of occurrences per acquired gloss)}} \\
        Random                      & 6.4 & 7.5 & 8.5 & 9.6 & 11.3 & 13.1 & 14.8 & 17.7 & 20.3 \\
        Entropy                     & 6.5 & 7.9 & 8.7 & 9.6 & 11.0 & 11.8 & 12.8 & 15.0 & 17.2 \\
        UNCAST                      & 5.7 & 6.5 & 7.2 & 7.9 & 8.9 & 9.8 & 10.8 & 13.0 & 15.1 \\
        Core-set                    & 6.4 & 7.5 & 8.2 & 9.0 & 10.5 & 11.8 & 12.9 & 14.9 & 17.0 \\
        Soft-Attn Core-set          & 6.4 & 8.1 & 9.5 & 10.8 & 13.6 & 15.4 & 17.1 & 20.0 & 22.8 \\
        \textbf{RAIDAL}             & 6.7 & 8.3 & 9.7 & 10.8 & 12.9 & 14.4 & 15.8 & 18.4 & 20.8 \\
        \bottomrule
    \end{tabular}
    }
\caption{Vocabulary statistics for CorrNet on PHOENIX-T across annotation budgets. Values are means over three independent replications.}
\label{tab:supp_pt_corrnet}
\end{table}

\begin{table}[H]
    \centering
    \small
    \setlength{\tabcolsep}{3pt}
    \resizebox{\linewidth}{!}{
    \begin{tabular}{l|cccccccc}
        \toprule
        \multirow{2}{*}{\textbf{Method}} & \multicolumn{8}{c}{\textbf{Annotation Budget in Frames}} \\
        \cmidrule{2-9}
        & \textbf{30k} & \textbf{40k} & \textbf{50k} & \textbf{60k} & \textbf{80k} & \textbf{100k} & \textbf{120k} & \textbf{160k} \\
        \midrule
        \multicolumn{9}{l}{\textit{(a) Unique Glosses Acquired}} \\
        Random              & 369 & 414 & 454 & 486 & 539 & 580 & 612 & 683 \\
        Entropy             & 385 & 440 & 495 & 550 & 639 & 723 & 800 & 905 \\
        UNCAST              & 401 & 464 & 521 & 574 & 668 & 740 & 814 & 910 \\
        Core-set            & 386 & 447 & 501 & 562 & 639 & 707 & 770 & 861 \\
        Soft-Attn Core-set  & 364 & 399 & 424 & 450 & 507 & 572 & 630 & 716 \\
        \textbf{RAIDAL (Ours)}     & 358 & 402 & 435 & 472 & 546 & 621 & 695 & 776 \\
        \midrule
        \multicolumn{9}{l}{\textit{(b) Vocabulary Coverage (\%)}} \\
        Random              & 30.0 & 33.6 & 36.9 & 39.5 & 43.8 & 47.1 & 49.7 & 55.5 \\
        Entropy             & 31.3 & 35.7 & 40.2 & 44.7 & 51.9 & 58.7 & 65.0 & 73.5 \\
        UNCAST              & 32.6 & 37.7 & 42.3 & 46.6 & 54.3 & 60.1 & 66.1 & 73.9 \\
        Core-set            & 31.4 & 36.3 & 40.7 & 45.7 & 51.9 & 57.4 & 62.6 & 69.9 \\
        Soft-Attn Core-set  & 29.6 & 32.4 & 34.4 & 36.6 & 41.2 & 46.5 & 51.2 & 58.2 \\
        \textbf{RAIDAL (Ours)}     & 29.1 & 32.7 & 35.3 & 38.3 & 44.4 & 50.4 & 56.5 & 63.0 \\
        \midrule
        \multicolumn{9}{l}{\textit{(c) Singletons (glosses appearing once)}} \\
        Random              & 149 & 158 & 172 & 179 & 191 & 203 & 211 & 230 \\
        Entropy             & 149 & 159 & 178 & 197 & 228 & 253 & 284 & 324 \\
        UNCAST              & 163 & 188 & 205 & 213 & 238 & 251 & 285 & 322 \\
        Core-set            & 151 & 168 & 182 & 200 & 230 & 251 & 273 & 305 \\
        Soft-Attn Core-set  & 140 & 144 & 144 & 154 & 171 & 188 & 213 & 235 \\
        \textbf{RAIDAL (Ours)}     & 136 & 150 & 158 & 170 & 192 & 227 & 255 & 276 \\
        \midrule
        \multicolumn{9}{l}{\textit{(d) Mean Gloss Frequency (mean number of occurrences per acquired gloss)}} \\
        Random              & 6.7 & 7.9 & 9.0 & 10.1 & 12.1 & 14.1 & 16.0 & 19.1 \\
        Entropy             & 6.7 & 8.0 & 9.1 & 9.9 & 11.5 & 12.7 & 13.6 & 15.8 \\
        UNCAST              & 6.1 & 7.0 & 7.6 & 8.3 & 9.5 & 10.8 & 11.7 & 14.0 \\
        Core-set            & 6.5 & 7.4 & 8.4 & 9.0 & 10.5 & 11.8 & 13.0 & 15.3 \\
        Soft-Attn Core-set  & 7.4 & 9.2 & 11.0 & 12.6 & 15.0 & 16.7 & 18.3 & 21.3 \\
        \textbf{RAIDAL (Ours)}     & 7.4 & 9.0 & 10.5 & 11.7 & 13.6 & 15.0 & 16.1 & 19.1 \\
        \bottomrule
    \end{tabular}
    }
\caption{Vocabulary statistics for Swin-MSTP on PHOENIX-2014 across annotation budgets. Values are means over three independent replications.}
\label{tab:supp_phoenix14_swin}
\end{table}

\begin{table}[H]
    \centering
    \small
    \setlength{\tabcolsep}{5pt}
    \resizebox{\linewidth}{!}{
    \begin{tabular}{l|cccccccc}
        \toprule
        \multirow{2}{*}{\textbf{Method}} & \multicolumn{8}{c}{\textbf{Annotation Budget in Frames}} \\
        \cmidrule{2-9}
        & \textbf{31k} & \textbf{41k} & \textbf{52k} & \textbf{62k} & \textbf{82k} & \textbf{103k} & \textbf{124k} & \textbf{165k} \\
        \midrule
        \multicolumn{9}{l}{\textit{(a) Unique Glosses Acquired}} \\
        Random                      & 329 & 371 & 406 & 435 & 490 & 528 & 560 & 623 \\
        Entropy                     & 349 & 392 & 438 & 495 & 586 & 671 & 748 & 842 \\
        UNCAST                      & 370 & 434 & 498 & 556 & 638 & 718 & 772 & 856 \\
        Core-set                    & 348 & 412 & 463 & 516 & 596 & 660 & 711 & 781 \\
        Soft-Attn Core-set          & 333 & 369 & 393 & 416 & 467 & 506 & 560 & 644 \\
        \textbf{RAIDAL}             & 332 & 375 & 408 & 444 & 497 & 561 & 622 & 717 \\
        \midrule
        \multicolumn{9}{l}{\textit{(b) Vocabulary Coverage (\%)}} \\
        Random                      & 30.3 & 34.2 & 37.4 & 40.1 & 45.1 & 48.7 & 51.6 & 57.4 \\
        Entropy                     & 32.2 & 36.2 & 40.4 & 45.6 & 54.0 & 61.8 & 68.9 & 77.6 \\
        UNCAST                      & 34.1 & 40.0 & 45.9 & 51.3 & 58.8 & 66.2 & 71.2 & 78.9 \\
        Core-set                    & 32.0 & 38.0 & 42.7 & 47.6 & 54.9 & 60.8 & 65.5 & 72.0 \\
        Soft-Attn Core-set          & 30.7 & 34.0 & 36.3 & 38.4 & 43.0 & 46.7 & 51.6 & 59.3 \\
        \textbf{RAIDAL}             & 30.6 & 34.6 & 37.6 & 41.0 & 45.8 & 51.7 & 57.4 & 66.1 \\
        \midrule
        \multicolumn{9}{l}{\textit{(c) Singletons (glosses appearing once)}} \\
        Random                      & 125 & 142 & 157 & 165 & 180 & 192 & 198 & 214 \\
        Entropy                     & 135 & 146 & 165 & 185 & 220 & 251 & 278 & 305 \\
        UNCAST                      & 150 & 179 & 208 & 233 & 252 & 278 & 293 & 306 \\
        Core-set                    & 140 & 161 & 177 & 203 & 232 & 252 & 267 & 286 \\
        Soft-Attn Core-set          & 129 & 143 & 146 & 152 & 162 & 171 & 193 & 228 \\
        \textbf{RAIDAL}             & 126 & 143 & 155 & 162 & 172 & 198 & 220 & 257 \\
        \midrule
        \multicolumn{9}{l}{\textit{(d) Mean Gloss Frequency (mean number of occurrences per acquired gloss)}} \\
        Random                      & 6.4 & 7.5 & 8.5 & 9.6 & 11.3 & 13.1 & 14.8 & 17.8 \\
        Entropy                     & 6.5 & 8.0 & 9.2 & 9.9 & 11.1 & 12.0 & 12.8 & 14.8 \\
        UNCAST                      & 5.7 & 6.5 & 7.1 & 7.6 & 8.8 & 9.7 & 10.8 & 13.0 \\
        Core-set                    & 6.2 & 7.0 & 7.8 & 8.4 & 9.7 & 10.8 & 11.9 & 14.3 \\
        Soft-Attn Core-set          & 6.7 & 8.4 & 10.2 & 11.8 & 14.2 & 16.5 & 17.8 & 20.5 \\
        \textbf{RAIDAL}             & 6.8 & 8.3 & 9.8 & 10.9 & 13.2 & 14.6 & 15.8 & 18.0 \\
        \bottomrule
    \end{tabular}
    }
\caption{Vocabulary statistics for Swin-MSTP on PHOENIX-T across annotation budgets. Values are means over three independent replications.}
\label{tab:supp_pt_swin}
\end{table}

\begin{table}[H]
    \centering
    \small
    \setlength{\tabcolsep}{5pt}
    \resizebox{\linewidth}{!}{
    \begin{tabular}{l|ccccc}
        \toprule
        \multirow{2}{*}{\textbf{Method}} &
        \multicolumn{5}{c}{\textbf{Annotation Budget in Frames}} \\
        \cmidrule{2-6}
        & \textbf{300k}
        & \textbf{400k}
        & \textbf{500k}
        & \textbf{600k}
        & \textbf{700k} \\
        \midrule
        \multicolumn{6}{l}{\textit{(a) Unique Glosses Acquired}} \\
        Random              & 625 & 649 & 660 & 668 & 672 \\
        Entropy             & 629 & 655 & 669 & 673 & 677 \\
        UNCAST              & 620 & 636 & 648 & 657 & 666 \\
        Core-set            & 629 & 657 & 671 & 676 & 677 \\
        Soft-Attn Core-set  & 630 & 648 & 657 & 666 & 670 \\
        \textbf{RAIDAL}     & 627 & 647 & 661 & 671 & 676 \\
        \midrule
        \multicolumn{6}{l}{\textit{(b) Vocabulary Coverage (\%)}} \\
        Random              & 91.9 & 95.4 & 97.1 & 98.2 & 98.8 \\
        Entropy             & 92.5 & 96.3 & 98.4 & 99.0 & 99.6 \\
        UNCAST              & 91.2 & 93.5 & 95.3 & 96.6 & 97.9 \\
        Core-set            & 92.5 & 96.6 & 98.7 & 99.4 & 99.6 \\
        Soft-Attn Core-set  & 92.6 & 95.3 & 96.6 & 97.9 & 98.5 \\
        \textbf{RAIDAL}     & 92.2 & 95.1 & 97.2 & 98.7 & 99.4 \\
        \midrule
        \multicolumn{6}{l}{\textit{(c) Singletons (glosses appearing once)}} \\
        Random              & 90 & 67 & 47 & 33 & 20 \\
        Entropy             & 85 & 50 & 29 & 13 & 9 \\
        UNCAST              & 93 & 66 & 52 & 40 & 28 \\
        Core-set            & 84 & 58 & 45 & 30 & 17 \\
        Soft-Attn Core-set  & 83 & 50 & 38 & 30 & 23 \\
        \textbf{RAIDAL}     & 94 & 58 & 41 & 28 & 17 \\
        \midrule
        \multicolumn{6}{l}{\textit{(d) Mean Gloss Frequency (mean number of occurrences per acquired gloss)}} \\
        Random              & 10.4 & 13.2 & 16.3 & 19.4 & 22.6 \\
        Entropy             & 12.4 & 17.2 & 21.5 & 25.8 & 29.7 \\
        UNCAST              & 10.4 & 13.4 & 16.2 & 19.1 & 21.9 \\
        Core-set            & 11.4 & 14.7 & 17.6 & 20.3 & 23.0 \\
        Soft-Attn Core-set  & 12.5 & 17.2 & 21.7 & 25.8 & 29.7 \\
        \textbf{RAIDAL}     & 12.4 & 17.2 & 21.7 & 26.0 & 30.0 \\
        \bottomrule
    \end{tabular}
    }
    \caption[Vocabulary statistics Swin-MSTP Isharah]
    {Vocabulary statistics for Swin-MSTP on Isharah across annotation
    budgets. All methods exceed 90\% vocabulary coverage by 300k frames. Values are means over three independent replications.}
    \label{tab:supp_isharah_swin}
\end{table}

\end{document}